%% file: BNPstreaming.tex
\documentclass[twoside]{article}
\usepackage[accepted]{aistats2015}

\usepackage{hyperref}
\usepackage{url}
\usepackage{graphicx}
\usepackage{amssymb}
\usepackage{epstopdf}
\usepackage{amsmath}

\usepackage{graphicx}
\usepackage{caption}
\usepackage{subcaption}
\usepackage[square,sort,comma,numbers]{natbib}
\usepackage{algorithm}
\usepackage{algorithmic}
\usepackage{mathtools}

\newcommand{\reals}{\mathbb{R}}

\newcommand{\E}{\mathbb{E}}
\newcommand{\KL}{\mathrm{KL}}

\newcommand{\DIR}{\mathrm{Dir}}
\newcommand{\MULT}{\mathrm{Mult}}
\newcommand{\POISP}{\mathrm{PP}}
\newcommand{\Levy}{L\'{e}vy }
\newcommand{\pred}{\mathrm{pr}}

\newcommand{\precap}{\vspace*{-0.25in}}
\newcommand{\postcap}{\vspace*{-0.1in}}

\usepackage{xcolor}

\DeclareMathOperator*{\argmin}{arg\,min}
\DeclareMathOperator*{\argmax}{arg\,max}

\begin{document}

\twocolumn[

\aistatstitle{Streaming Variational Inference for Bayesian Nonparametric Mixture Models}

\aistatsauthor{Alex Tank \And Nicholas J. Foti \And Emily B. Fox }

\aistatsaddress{ University of Washington \\ Department of Statistics \And University of Washington \\ Department of Statistics \And University of Washington \\ Department of Statistics } ]

%
%
%

\begin{abstract}
	\vspace{-0.1in}
In theory, Bayesian nonparametric (BNP) models are well suited to streaming data scenarios due to their ability
to adapt model complexity with the observed data.  Unfortunately, such benefits have not been fully realized in practice; existing
inference algorithms are either not applicable to streaming applications or not extensible
to BNP models.  For the special case of Dirichlet processes, streaming inference has been considered.  However, there is growing interest in more flexible BNP models building on the class of normalized random measures (NRMs).  We work within this general framework and present a streaming variational inference algorithm for NRM mixture models. Our algorithm is based on assumed density filtering (ADF), leading straightforwardly to expectation propagation (EP) for large-scale batch inference as well.  
We demonstrate the efficacy of the algorithm on clustering documents in large, streaming text corpora.
\end{abstract}

\vspace{-0.15in}
\input{intro}
\input{background}

\input{algorithm}

\input{experiments}
\input{discussion}

\bibliography{BNPstreaming}
\bibliographystyle{unsrt}

\setcounter{section}{0}
\setcounter{figure}{0}
\setcounter{table}{0}
\setcounter{equation}{0}
\setcounter{footnote}{0}

\twocolumn[

\aistatstitle{Supplemental Information:  Streaming Variational Inference for Bayesian Nonparametric Mixture Models}]



\input{supplementADF}
\input{nrm_pred_general}

\input{supplementEP}

\input{experiments_supp}

\end{document}

%% file: intro.tex
\section{Introduction}
Often, data arrive sequentially in time and we are tasked with performing unsupervised learning as the data stream in, without revisiting past data.
For example, consider the task of assigning a topic to a news article based on a history of previously assigned documents.  The articles arrive daily---or more frequently---with no bound on the total
number in the corpus. In clustering such \textit{streaming} data, Bayesian nonparametric (BNP) models are natural as they allow the
number of clusters to grow as data arrive.  A challenge, however, is that it is infeasible to store the past cluster assignments, and instead inference algorithms must rely solely on summary statistics of these variables.

Stochastic variational inference (SVI)~\cite{Hoffman:Blei:Wang:Paisley:2013} has become a popular method for scaling posterior inference in Bayesian latent variable models. Although SVI has been extended to BNP models, SVI requires specifying the size of the data set a priori, an inappropriate assumption for streaming data.
In contrast, streaming variational Bayes (SVB)~\cite{Broderick:2013} handles unbounded data sets by exploiting the sequential nature of Bayes theorem to recursively update
an approximation of the posterior.
Specifically, the variational approximation of the current posterior becomes the prior when considering new observations. 
While SVB
is appropriate for parametric models, it does not
directly generalize to the BNP setting that is essential for
streaming data.

For BNP models, streaming inference has been limited to algorithms
hand-tailored to specific models.  For example, a streaming variational
inference algorithm for Dirichlet process (DP) mixture models was recently
proposed based on heuristic approximations to the Chinese restaurant process
(CRP) predictive distribution associated with the DP~\cite{Lin:2013}.

We seek a method for streaming inference in BNP models that is more generally
extensible.  We are motivated by the recent focus on a broader class of BNP priors---\emph{normalized random measures} (NRMs)---that
enable greater control of various properties than the DP permits.  For example,
in clustering tasks, there is interest in having flexibility in the
distribution of cluster sizes.  
Throughout the paper, we focus on the specific case of the normalized generalized gamma process (NGGP), though our methods are more general.
Recently, NGGP mixture models have been shown to outperform
the DP~\cite{Barrios:Lijoi:Nieto-Barajas:Prunster:2013, Favaro:Teh:2013}, 
but inference has relied on Markov chain Monte Carlo (MCMC).  
Due to the limitations of MCMC, such demonstrations have been limited to small
data sets.  Importantly, NGGPs and the DP differ mainly in their
asymptotic scaling properties and the use of NGGPs may be more appropriate in
large data sets where the logarithmic cluster growth rate of the DP is not
appropriate.


To address the challenge of streaming inference in NRM mixture models, we develop a variational algorithm based on assumed density filtering (ADF)~\cite{Minka:2001}.  Our algorithm uses infinite-dimensional approximations to the
mixture model posterior and allows general BNP
predictive distributions to be used by leveraging an auxiliary variable representation.  As a byproduct of the ADF construction, a multi-pass variant straightforwardly yields an expectation propagation (EP) algorithm for batch inference in BNP models.  This provides a new approach to scalable BNP batch inference.

In the special case of DPs, our algorithm reduces to that of~\cite{Lin:2013}. As such, our framework forms a theoretically justified and general-purpose scheme for BNP
streaming inference, encompassing previous heuristic and model-specific approaches, and with a structure that enables insight into BNP inference via EP.
%

We demonstrate our algorithm on clustering documents from text corpora using an
NRM mixture model based on the NGGP~\cite{Favaro:Teh:2013}.  After a single pass through a modest-sized data set, our
streaming variational inference algorithm achieves performance nearly on par
with that of a batch sampling-based algorithm that iterates through the data
set hundreds of times.  We likewise examine a New York Times corpus
of 300,000 documents to which the batch algorithm simply does not scale (nor would it be applicable in a truly streaming setting).  In
these experiments, we justify the importance of considering the flexible class of NRM-based models.  Our work represents the first application of non-DP-based NRMs to such large-scale applications.

%% file: background.tex
\section{Background}
%
%
\subsection{Completely Random Measures}

A \textit{completely random measure} (CRM)~\cite{Kingman:1967} is a
distribution over measures $G$ on $\Theta$ 
such that for disjoint
$A_k \subset \Theta$, $G(A_k)$ are independent random variables and
%
\begin{align}
G = \sum_{k=1}^\infty \pi_k \delta_{\theta_k}.
\end{align}
The masses $\pi_k$ and locations $\theta_k$ are 
characterized by a Poisson process on $\Theta \times \reals_+$ with
\Levy measure $\mu(d\theta, d\pi)$~\cite{Kingman:1967,Kingman:1993}.
We restrict our attention to \textit{homogeneous} CRMs where
$\mu(d\theta, d\pi) = H_0(d\theta) \lambda(d\pi)$, a common assumption
in the literature ~\cite{Ferguson:1973, Hjort:1990, James:Lijoi:Prunster:2009}. 
 We denote
a draw from a homogeneous CRM as
\begin{align}
G \sim \mathrm{CRM}(\lambda, H_0).
\end{align}
The total mass $T=G(\Theta)=\sum_{k=1}^\infty \pi_k$ is almost surely
finite~\cite{Regazzini:2003}.  However, since $T\neq 1$ in general,
CRMs cannot directly be used as priors for mixture models.

\subsection{Normalized Random Measures}
\label{sec:nrm}

One can normalize a CRM by its finite total mass to construct a BNP prior for mixture models. 
Specifically, define
the \textit{normalized random measure} (NRM) $P = \sum_{k=1}^\infty
\frac{\pi_k}{T} \delta_{\theta_k}$.  The Dirichlet process (DP) is an NRM which arises from
normalizing the masses of a gamma process~\cite{Ferguson:1973}.  However, more
flexible NRMs can be constructed by starting with different CRMs.

In the mixture model setting, we observe data
$\{x_i \in \reals^d\}$ with $x_i$ 
generated from mixture component
$\theta^{z_{i}}$. Here, we assume the assignment variables, $z_i$, are 1-of-$K$ coded so that $\sum_k z_{ik} = 1$ and $z_{ik} = 1$
implies that observation $i$ is assigned to component $\theta_k$ via $\theta^{z_i}$. The resulting NRM mixture model can be written as:
\begin{equation}
\begin{aligned}
    G \; | \; \lambda, H_0 &\sim \mathrm{CRM}(\lambda, H_0) \\
    z_i \; | \; G &\sim \sum_{k=1}^\infty \frac{\pi_k}{T} \delta_k\\
    x_i \; | \; z_i, \theta &\sim F(x_i | \theta^{z_i}),
\end{aligned}
\label{eqn:nrmmix}
\end{equation}
where $F(\cdot|\cdot)$ is an observation model. 

For our running example of the \textit{normalized generalized gamma process} (NGGP), the GGP L\'{e}vy
measure is
\begin{equation}
    \lambda(d\pi) = \frac{a}{\Gamma(1-\sigma)} \pi^{-\sigma - 1}e^{-\tau \pi}d\pi,
\end{equation}
where $\tau \in [0,\infty)$, $a \in (0, \infty)$, and $\sigma \in [0,1)$.
Notable special cases of the NGGP are $\sigma=0$, where we obtain the DP,
and $\sigma=0.5$, where we obtain the normalized inverse-Gaussian (IG) process.  The NGGP
with $\sigma \neq 0$ provides greater
control over model properties, such as the distribution of cluster sizes~\cite{Barrios:Lijoi:Nieto-Barajas:Prunster:2013}.

For any NRM mixture model, by introducing an auxiliary variable $U_n \sim
\Gamma(n, T)$, we can integrate out the NRM $P$ and define a partial urn
scheme~\cite{Favaro:Teh:2013,James:Lijoi:Prunster:2009}.  In the case of the
NGGP we have:
\begin{equation}
   \label{eqn:nrmurn}
   \hspace{-0.25em}
   p(z_{{(n+1)}k} |U_n, z_{1:n}) {\propto}
    \left\{\begin{array}{ll}
    \displaystyle n_k - \sigma, \,\,\,\,\,\,\,\,\,\,\,\,\,\,  k \leq K \\
    a(U_n + \tau)^{\sigma}, \,\, k = K+1,
\end{array}\right.
\end{equation}
where $K$ is the number of instantiated clusters in $z_{1:n}$.  When $\sigma=0$,
Eq.~\eqref{eqn:nrmurn} reduces to the well known Chinese restaurant process (CRP)
corresponding to the DP.  The posterior distribution of $U_n$ is given
by~\cite{James:Lijoi:Prunster:2009}:
 \begin{equation}
 \label{eqn:u}
 p(U_n|z_{1:n}) \propto \frac{U^{n}_{n}}{(U_{n} + \tau)^{n - a K}} e^{-\frac{a}{\sigma}(U_{n} + \tau)^{\sigma}}.
\end{equation}
Together, Eqs.~\eqref{eqn:nrmurn} and~\eqref{eqn:u} can be used to define MCMC
samplers for NGGP mixture models~\cite{Favaro:Teh:2013,Griffin:Walker:2011}; our
streaming algorithm also exploits the use of $U_n$.

\subsection{Assumed Density Filtering}
\label{sec:ADF}

Assumed density filtering (ADF) was first developed as a sequential procedure for inference in dynamic models that iteratively projects
an intractable distribution onto a simpler family of distributions. Let
$z_{1:n} = (z_1, z_2, \ldots, z_n)$ be a sequence of random variables with
joint distribution $p_n(z_{1:n})$.  We can
write the joint distribution as a product of factors,
\begin{equation}
p_n(z_{1:n}) \propto \prod_{i=1}^n f_i(z_{1:i}).
\label{eq:ADFfact}
\end{equation}
ADF approximates the sequence of distributions $p_{n}(z_{1:n})$
with a sequence $\hat{q}_n(z_{1:n}) \in \mathcal{Q}_n$, where
$\mathcal{Q}_n$ is a family of simpler distributions. Based on the current
$\hat{q}_n(z_{1:n})$, the approximation to $p_{n+1}(z_{1:n+1})$ is formed as
follows.  The $(n+1)$st factor is incorporated to 
form $\hat{p}_{n+1}(z_{1:n}) \stackrel{\triangle}{\propto} f_{n+1}(z_{1:n+1}) \hat{q}_n(z_{1:n})$,
which is then projected onto
$\mathcal{Q}_{n + 1}$ by minimizing the KL divergence:
\begin{multline}
\hat{q}_{n+1}(z_{1:n+1}) = \\
\argmin_{q_{n+1} \in \mathcal{Q}_{n+1}}
\KL \Big ( \hat{p}_{n+1}(z_{1:n+1}) || q_{n+1}(z_{1:n+1}) \Big ). 
\label{eqn:KLproj}
\end{multline}
When $\mathcal{Q}_n$ factorizes as $q_n(z_{1:n}) = \prod_{i=1}^n q_n(z_i)$,
the optimal distribution for each factor is given by the marginal distribution,
$\hat{q}_{n+1}(z_{i}) \propto \int  f_{n+1}(z_{1:n + 1}) \hat{q}_n(z_{1:n}) dz_{\backslash i}$, where $z_{\backslash i}$ denotes the set $\{z_j, j\neq i\}$. The tractability of this integral for certain families of factors $f_n$ and $\hat{q}_n$ motivates ADF, and in particular, the recursive projection onto $\{\mathcal{Q}_n\}$.

\subsection{Expectation Propagation}
\label{sec:EP}

ADF can be generalized to perform batch inference in static models resulting
in the well known expectation propagation (EP) algorithm~\cite{Minka:2001}.
In EP, one approximates an intractable, factorized distribution over a
fixed set of model parameters, $\theta$, with a tractable distribution, $q \in
\mathcal{Q}$. In place of Eq.~\eqref{eq:ADFfact}, we have
\begin{equation}
p(\theta) \propto \prod_{i = 1}^{n} f_i(\theta).
\end{equation}
An EP iteration begins with both a posterior approximation, $\hat{q}(\theta)$,
and stored local contributions, $\bar{q}_j(\theta)$, associated with each
factor $f_j(\theta)$. To refine the posterior approximation, a local
contribution is removed to form a normalized approximation to the remaining
$n-1$ factors, $\hat{q}_{\setminus j}(\theta) \propto \frac{q_{\theta)}}{\bar{q}_j(\theta)}$.
As in ADF, the $j$th factor is then appended to the approximation $\hat{q}_{\setminus j}$ and projected back onto $\mathcal{Q}$ to obtain a refined $\hat{q}(\theta)$:
\begin{equation} \label{epupdate}
    \hat{q}(\theta) = \argmin_{q \in \mathcal{Q}}\KL \Big ( \hat{p}(\theta)
    \propto f_{j}(\theta) \hat{q}_{\setminus j}(\theta) \Big | \Big | q(\theta) \Big ).
\end{equation}
The $j$th local contribution is then updated to
\begin{equation}
\bar{q}_j(\theta) \propto \frac{\hat{q}(\theta)}{\hat{q}_{\setminus j} (\theta)}.
\end{equation}
When $\hat{q}, \bar{q}_j, \hat{q}_{\setminus j}$ are in the exponential family with the same type of
sufficient statistics, $\hat{\nu}, \bar{\nu}_j, \hat{\nu}_{\setminus j} \in \mathbb{R}^{m}$,
respectively, then $\bar{\nu}_j = 
\hat{\nu} - \hat{\nu}_{\setminus j}$. This process of removing local statistics from the approximation, adding in the respective factor, and re-projecting onto $\mathcal{Q}$ is repeated for all factors until convergence. 

The link between ADF and EP, comparing Eqs.~\eqref{eqn:KLproj} and \eqref{epupdate}, allows us to extend our streaming
BNP algorithm of Sec.~\ref{sec:ADFNRM} to EP for batch inference (Sec.~\ref{sec:EPNRM}). EP is easily parallelized \cite{Gelman:2014}, allowing these methods to scale to massive batch data sets, though we leave the parallel extension of our method to future work. 

%% file: algorithm.tex
\section{Streaming Variational Inference for BNP Mixture Models}
\label{sec:ADFNRM}

We now turn to deriving a streaming inference algorithm for the NRM mixture
model of Eq.~\eqref{eqn:nrmmix}.  Here, our goal is joint inference of the
growing set of local cluster indicators, $z_{1:n}$, and the static set of
global cluster parameters, $\theta = \{\theta_k\}_{k=1}^\infty$.  The method is 
derived from the ADF algorithm of Sec.~\ref{sec:ADF} and boils
down to: (1) a local update of cluster soft assignments for the current data point and (2) a global update of cluster variational parameters. The local update follows directly from ADF. Embedded in this step is computing the NRM predictive probability on cluster assignments, for which we use the auxiliary variable representation of Eq.~\eqref{eqn:nrmurn} combined with an additional variational approximation to compute an intractable integral.  For computational tractability, the global step uses an approximation similar to that proposed in \cite{Wang:Blei:2012}, though an exact ADF update is possible.


To start, note that the posterior for the first $n$ assignments, $z_{1:n}$, and
cluster parameters, $\theta$, factorizes as:
\begin{align} 
p_n(z_{1:n}, \theta| x_{1:n}) &\propto p(x_n| z_n, \theta) p(z_n | z_{1:n - 1}) \label{eqn:seqposterior} \\
& \hspace{0.1em}\times p(z_{1:n-1}, \theta|x_{1:n-1}) \nonumber \\
&\propto p(\theta) \prod_{i = 1}^{n} p(x_i | z_i, \theta) p(z_i| z_{1:i-1}).
\label{eqn:factposterior}
\end{align}
Eq.~\eqref{eqn:seqposterior} emphasizes the sequential decomposition of the posterior while 
Eq.~(\ref{eqn:factposterior}) concretely links our derivation with ADF. 
We set the first factor to $p(x_1|z_1,\theta) p(z_1) \prod_{k = 1}^{\infty}
p(\theta_k)$, where $\mbox{$p(z_{11} = 1) = 1$}$ so that $p(x_1|z_1,\theta) p(z_1) = p(x_1|\theta_1) p(z_1)$. For $i > 1$, we define
$p(z_i | z_{1:i - 1})$ as the $i$th \emph{predictive} factor and $p(x_i| z_i, \theta)$ as the $i$th \emph{likelihood} factor. 
We then apply ADF to this sequence of factors in Eq.~\eqref{eqn:factposterior} to obtain a sequence of factorized variational approximations of the form $\hat{q}_{n}(z_{1:n}, \theta) =  \prod_{k = 1}^{\infty}\hat{q}_{n}(\theta_k)  \prod_{i = 1}^{n} \hat{q}_{n}(z_i)$. Since the first factor takes this factorized form, we have
$\hat{q}_1(z_1,\theta) \propto  p(z_1) p(x_1|\theta_1) p(\theta_1)
\prod_{k = 2}^{\infty}p(\theta_k)$; algorithmically we only update $\hat{q}_1(z_1)$ and $\hat{q}_1(\theta_1)$. For subsequent factors,
assume the posterior $p(z_{1:n-1}, \theta| x_{1:n-1})$ is approximated by a factorized
$\hat{q}_{n - 1}(z_{1:n-1}, \theta)$. For $n > 2$, we add $p(z_n | z_{1:n - 1})$, perform an ADF step, and then add $p(x_n| z_n, \theta)$ and perform another ADF step.

\paragraph{Predictive factors} 

To approximate the posterior after adding the $p(z_n | z_{1:n - 1})$ factor, 
we use Eq.~\eqref{eqn:KLproj}:
\vspace{-8 pt}
\begin{align}
    & q^{\pred}(z_{1:n}, \theta) = \argmin_{q_n \in \mathcal{Q}_n}
    \KL \Big ( \hat{p}_n(z_{1:n},\theta | x_{1:n -1}) || q_n(z_{1:n}, \theta) \Big ). \nonumber
\end{align}
\vspace{-12 pt}

where $\hat{p}_n(z_{1:n},\theta | x_{1:n - 1}) \stackrel{\triangle}{\propto}
    p(z_n | z_{1:n - 1})
    \hat{q}_{n-1}(z_{1:n-1}, \theta)$ is the \emph{propagated} variational distribution and $q^{\pred}$ its projection back to $Q_n$.  
For $i < n$, the optimal approximation for the local variables, $z_{i}$, is $q^{\pred}(z_i) = \hat{q}_{n-1}(z_i)$, while for the $n$th local variable we have
    \vspace{-5 pt}
\begin{align}
q^{\pred}(z_n) = \sum_{z_{1:n-1}} p(z_n|z_{1:n-1}) \prod_{i = 1}^{n-1}
\hat{q}_{n-1}(z_i). \label{eqn:qpred1}
\end{align}
\vspace{-15 pt}

The combinatorial sum over $z_{1:n-1}$ embedded in evaluating $q^{\pred}(z_n)$ appears to be a daunting barrier to efficient streaming inference.  However, as we show in
Sec.~\ref{sec:pred_ggp}, for the models we consider the resulting $q^{\pred}$
can be written in terms of sums of local soft assignments, $\sum_{i=1}^{n-1}
\hat{q}_{n-1}(z_i)$.  Since these past soft assignments remain unchanged, the sum---instead of past assignment histories---can be stored as a sufficient
statistic. Furthermore, since
$p(z_n|z_{1:n-1})$ places mass on $z_n$ taking a previously unseen component,
the approximation $q^{\pred}(z_n)$ inherits this ability and allows our
algorithm to introduce new components when needed. This is a crucial feature of
our approach that enables our approximate inference scheme to maintain the
benefits of nonparametric modeling, and is in contrast to approaches based on
truncations to the underlying NRM or on heuristics for creating
new clusters. 

Since the predictive factor does not depend on $\theta$, the approximation for $\theta$ is retained: $q^{\pred}(\theta_j) = \hat{q}_{n-1}(\theta_j)$.

\paragraph{Likelihood factors}
We apply Eq.~\eqref{eqn:KLproj} to obtain the approximation after adding the $p(x_n|z_n,\theta)$ factor
\vspace{-5 pt}
\begin{align}
    \hat{q}_n(z_{1:n}, \theta) = \argmin_{q_n \in \mathcal{Q}_n}
    \KL \Big ( \hat{p}_n(z_{1:n},\theta | x_{1:n}) || q_n(z_{1:n}, \theta) \Big ),\nonumber
\end{align}
\vspace{-15 pt}

where  $\hat{p}_n(z_{1:n},\theta | x_{1:n}) \stackrel{\triangle}{\propto} p(x_n| z_n, \theta) q^{\pred}(z_{1:n}, \theta)$ is the \emph{updated} variational distribution. Projecting back to $Q_n$, for $i < n$ 
 the optimal
distributions for the $z_i$ are retained: $\hat{q}_n(z_i) =\hat{q}_{n-1}(z_i)$. For $z_{n}$, we have
\begin{align}
    \label{eqn:localup}
    \hat{q}_n(z_{nk}) &\propto q^{\pred}(z_{nk}) \int p(x_n| z_{nk}, \theta)
    \hat{q}_{n-1}(\theta_k) d\theta_k ,
\end{align}
where $q^{\pred}(z_{nk})$ mirrors the role of the predictive rule,
$p(z_n|z_{1:n-1})$, when assignments are fully observed. We consider conjugate exponential family models 
so that $\hat{q}_{n-1}(\theta_k)$ is in the same family as $p(\theta_k)$, allowing the integral in Eq.~\eqref{eqn:localup} to be given in closed form. 

The update in Eq.~\eqref{eqn:localup} has appeared previously in both batch~\cite{Wang:Blei:2012}
and streaming~\cite{Lin:2013} inference algorithms for DP
mixtures (without being derived from the ADF framework).
In the batch case, $q^\pred(z_n)$ was evaluated by sampling, and in the latter
case a heuristic approximation was used.  We instead use a principled
variational approximation to evaluate Eq.~\eqref{eqn:localup}, which extends to a
large class of NRMs. See Sec.~\ref{sec:pred_ggp}.


As in EP~\cite{Minka:2001}, the optimal update for the global parameters, $\theta_k$, after addition of the likelihood factor is proportional
to the marginal:
\vspace{-5 pt}
\begin{align}
    \label{eqn:thetaup}
    \hat{q}_n(\theta_k) \propto
            \sum_{z_{1:n}} \int \hspace{-0.2em} p(x_n| z_n, \theta) q^{\pred}(z_{1:n}, \theta) d \theta_{\backslash k}.
\end{align}
\vspace{-15 pt}

Eq.~\eqref{eqn:thetaup} is often intractable so we
use the conjugate variational Bayes update for $\theta_k$ as in~\cite{Wang:Blei:2012}, giving:
\begin{align}
    \log \hat{q}_n(\theta_k) &\approx \E_{\theta_{\backslash k}, z_n} \log [ p(x_n| z_n, \theta) \hat{q}_{n-1} (\theta)] + C,
\end{align}
where $C$ is a constant. See the Supplement for details.  The expectation is
with respect to the
distributions $\hat{q}_n(z_{n})$ and $\hat{q}_n(\theta_{\backslash k}) = \prod_{j \neq k} \hat{q}_{n}(\theta_j)$.
This implies that
\begin{align}
    \label{eqn:mfupdate}
    \log \hat{q}_n(\theta_k) \approx \hat{q}_n(z_{nk})\log p(x_n|z_{nk},\theta) 
    + \log \hat{q}_{n-1}(\theta_k) + C'.
\end{align}
\vspace{-15 pt}

For the conjugate models we consider,
Eq.~\eqref{eqn:mfupdate} leads to tractable updates. Our streaming
algorithm, which we refer to as \textit{ADF-NRM}, proceeds at each step by first computing the local update in
Eq.~\eqref{eqn:localup}, and then the global update in
Eq.~\eqref{eqn:mfupdate}.  See Alg.~\ref{alg:adfnrm}.



\subsection{Predictive Rule for NGGPs} \label{sec:pred_ggp}

A key part of the streaming algorithm is efficiently computing $q^{\pred}(z_n)$.
When a DP prior is used, $q^{\pred}(z_n)$ admits a simple form similar to the
CRP:
\begin{equation}
\label{eqn:qpred_dp}
q^{\pred}(z_{nk}) {\propto} 
    \left\{\begin{array}{ll}
            \displaystyle \sum_{i = 1}^{n-1} \hat{q}_i(z_{ik}), & k \leq K_{n-1} \\
            a, & k = K_{n-1}+1,
    \end{array}\right.
\end{equation}
where $K_{n-1}$ is the number of considered components in $x_{1:n-1}$ (see Sec.~\ref{sec:coping}).
Unfortunately, NRMs do not admit such a straightforward expression for 
$q^\pred(z_n)$ since in general $p(z_n|z_{1:n-1})$ is not known in closed form and for NGGPs it is given by a computationally demanding and numerically
unstable expression \cite{Lijoi:2007} unsuitable for large, streaming data.


Instead, as in Eq.~\eqref{eqn:nrmurn}, we can
introduce an
auxiliary variable, $U_n$, to obtain a tractable variational approximation
for NRMs, as detailed in the Supplement. We focus on the popular case of the NGGP here.

We rewrite $q^{\pred}(z_n)$ in terms of
$U_{n-1}$ and the
unnormalized masses, $\pi$, and integrate over these variables:
\begin{align}
    \label{eqn:nrmpred}
    q^{\pred}(z_n) = &\sum_{z_{1:n-1}} \hspace{-0.5em} \iint \hspace{-0.2em}
    \Bigg[ p(z_n|\pi) p(\pi|U_{n-1}, z_{1:n-1}) \\
        &\times p(U_{n-1}|z_{1:n-1}) \prod_{i = 1}^{n-1} \hat{q}_{n-1}(z_i)
    \Bigg ] dU_{n-1} d\pi. \nonumber
\end{align}
The term $p(\pi|U_{n}, z_{1:n-1})$ is stated in the Supplement and
$p(U_{n-1}|z_{1:n-1})$ is shown in Eq.~\eqref{eqn:u}.
The random measure $\pi$ consists of a set of instantiated atoms,
$\pi_1,\ldots,\pi_K$, and a Poisson process $\pi^*$ representing the remaining
mass.
Since the integral in
Eq.~\eqref{eqn:nrmpred} is intractable, we introduce a partially factorized
approximation:
$p(\pi|U_{n - 1}, z_{1:n-1}) p(U_{n -1}|z_{1:n-1}) \approx q(\pi|U_{n-1})
q(U_{n-1}) \in \mathcal{Q}_{\pi \times U}$
and solve
\begin{equation}
\begin{aligned}
    \argmin_{q \in \mathcal{Q}_{\pi \times U}} \KL \Big ( &q(\pi|U_{n-1}) q(U_{n-1}) \hat{q}(z_{1:n-1}) || \\
    &p(\pi|U_{n - 1}, z_{1:n-1}) p(U_{n -1}|z_{1:n-1}) \hat{q}(z_{1:n-1}) \Big ). \nonumber
\end{aligned}
\end{equation}
 The optimal distributions are given by:
\begin{align}
    q(U_{n - 1}) &\propto e^{-\frac{a}{\sigma}(U_{n - 1} + \tau)^{\sigma}}
\frac{U^{n-1}_{n - 1}}{(U_{n-1} + \tau)^{n - 1 - a \E_{\hat{q}}[K_{n
- 1}]}} \label{eqn:u_var} \\
q(\pi_k|U_{n-1}) &\propto \pi_k^{\E_{\hat{q}}[n_k]} e^{-U_{n-1} \pi_k} \lambda(d\pi_k),
\end{align}
where $\E_{\hat{q}_{n-1}}[K_{n - 1}]$ is the expected number of
clusters observed so far, which can be recursively computed as described in the Supplement, and $\E_{\hat{q}_{n-1}}[n_k] = \sum_{i = 1}^{n-1}
\hat{q}_{n-1}(z_{ik})$ is the expected number of assignments to component $k$.
The variational distribution of $\pi^*$ is a Poisson process with tilted \Levy
measure $e^{U_{n-1}\pi} \lambda(d\pi)$.
As detailed in the Supplement, using these variational approximations in Eq.~\eqref{eqn:nrmpred} combined with a delta function approximation to $q(U_{m-1})$ yields:
\begin{equation}\label{eqn:nrmpredapprox}
    q^{\pred}(z_{nk}) {\propto} 
\left\{\begin{array}{ll}
        \displaystyle \max \left(\sum_{i = 1}^{n-1} \hat{q}_i(z_{ik}) - \sigma, 0 \right),  k \leq
        K_{n-1} \\
        a(\hat{U}_{n - 1} + \tau)^{\sigma},  \,\,\,\,\,\,\,\,\,\,\,\,\,\,\,\,\,\,\,\  k = K_{n-1}+ 1,
\end{array}\right.
\end{equation}
where $\hat{U}_{n - 1} = \argmax q(U_{n-1})$. For the DP ($\sigma = 0$), 
Eq.~\eqref{eqn:nrmpredapprox} reduces to Eq.~\eqref{eqn:qpred_dp} and the
resulting algorithm reduces to that of \cite{Lin:2013}. Note the differences
between Eqs.~\eqref{eqn:nrmpredapprox} and \eqref{eqn:nrmurn} and between
Eqs.~\eqref{eqn:u_var} and \eqref{eqn:u}. In both cases, hard assignments are replaced by soft assignments.  As previously noted, the sum of these past soft assignments serve as sufficient statistics, and since they do not change between iterations, can be stored in place of individual assignments.
%
Furthermore, the recursive computation of $\E_{\hat{q}_{n-1}}[K_{n - 1}]$ in Eq.~\eqref{eqn:u_var}
allows past assignments to be discarded.

\begin{algorithm}[t!]
\begin{algorithmic}
\STATE {Initialize: $K = 1, S_1 = 1$}
\STATE {$\hat{q}_1(\theta_1) \propto p(x_1|\theta_1) p(\theta_1)$, $\hat{q}_1(z_{11}) = 1$}
\FOR{$n = 1$ to $\infty$  }
	\STATE {$\hat{U}_n = \argmax q(U_n) \mbox{ with $q(U_n)$ in   Eq. \eqref{eqn:u_var}} $}
	\FOR{$k = 1$ to K}
		\STATE $q^{\pred}(z_{nk}) \propto \max(S_k - \sigma,0)$
		\STATE $\hat{q}_n(z_{nk}) \propto q^{\pred}(z_{nk}) \int
        p(x_n|z_{nk},\theta_k) \hat{q}_{n-1}(\theta_k) d\theta_k$
	\ENDFOR
	\STATE $q^{\pred}(z_{n, K + 1}) \propto a(\hat{U}_n + \tau)^{\sigma}$
	\STATE $\hat{q}_n(z_{n, K + 1}) \propto q^{\pred}(z_{n, K + 1})$
	\STATE $\,\,\,\,\,\,\,\,\,\,\,\,\,\,\,\,\,\,\,\,\,\,\,\,\,\,\,\,\,\,\,\,\,\,\,\,\,\,\,\,\,\,\,
    \hspace{-1.1em} \times \int \hspace{-0.2em}
        p(x_n|z_{n, K+1},\theta) p(\theta_{K+1}) d\theta_{K+1}$
    \STATE{normalize $\hat{q}_{n}(z_{n,1:K+1})$}
	\IF {$\hat{q}_n(z_{n, K + 1}) > \epsilon$} 
		\STATE $S_{K+1} = 0$, $\hat{q}_{n-1}(\theta_{K+1}) = p(\theta_{K+1}), K
        = K + 1$
        
	\ELSE 
		\STATE{normalize $\hat{q}_{n}(z_{n,1:K})$}
	\ENDIF
	\FOR{$k = 1$ to $K$}
		\STATE $\hat{q}_n(\theta_k) \propto p(x_n|z_{nk},\theta_k)^{\hat{q}_n(z_{nk})} \hat{q}_{n - 1}(\theta_k)$
		\STATE $S_k = S_k + \hat{q}_n(z_{nk})$
	\ENDFOR
\ENDFOR
\end{algorithmic}
\caption{ADF for NRM mixture models}
    \label{alg:adfnrm}
\end{algorithm}

\subsection{Computational Complexity}

Due to the streaming nature of the ADF-NRM algorithm, we analyze the
per-observation complexity.  As seen in Alg.~\ref{alg:adfnrm},
for each observation we compute a finite dimensional probability vector with
$K_n+1$ elements, which is $O(K_n)$.  Additionally, we need to compute
$\hat{U}_n$ via numerical optimization of $q(U_n)$, which is a univariate and
unimodal function so can be maximized efficiently with complexity denoted
$O(\mathcal{U})$.  Thus, the per-iteration complexity of ADF-NRM
is $O(K_n + \mathcal{U})$. In practice the runtime is dominated
by the $O(K_n)$ term due to the NGGP introducing many clusters; the optimization
of $\hat{U}_n$ terminates in a few iterations (independent of $K_n$) and so
does not limit the scalability.  It is known that $\E[K_n] \simeq a \log n$ for the DP and
follows a power-law with index $\sigma \in (0,1)$ for the
NGGP~\cite{Favaro:2012}.  This implies that for large $n$ the complexity of
ADF-NRM with a NGGP is larger than that with a DP, but is sub-linear in $n$,
remaing computationally feasible.  Of course, a posteriori $K_n$ can grow
much more slowly when the data has a compact representation.

\subsection{Efficiently Coping with New Clusters}
\label{sec:coping}

While the probability that a data point belongs to a new cluster,
$\hat{q}_n(z_{n,K+1})$, is always greater than zero, it is computationally
infeasible to introduce a new component at each iteration since the per
iteration complexity of ADF-NRM is $O(K_n)$.  In practice, new components are
added only if $\hat{q}_{n}(z_{n,K_n+1}) > \epsilon$ for $\epsilon \geq \sigma$ a threshold.  The restriction $\epsilon \geq \sigma$ is natural: if $\hat{q}_{n}(z_{n,K_n+1}) < \sigma$ then $K_{n}+1$ will be assigned zero prior probability at step $n + 1$ in Eq. \eqref{eqn:nrmpredapprox} and will be effectively removed.  The threshold parameter explicitly
controls the trade off between accuracy and speed; a larger threshold
introduces fewer clusters leading to a worse variational approximation but
faster run times. One can view our thresholding as an adaptive truncation of the posterior, in contrast to the common approach of truncating the component prior. 

During execution of ADF-NRM and EP-NRM of Sec.~\ref{sec:EPNRM}, redundant clusters can be created due to
the order of observations processed.  As in~\cite{Lin:2013}, we
introduce merge steps to combine distinct clusters that explain similar
observations. Since a benefit of the NGGP over the DP is the addition of many small but important clusters (see Sec.~\ref{sec:experiments}), we found that frequent merging degrades predictive performance of NGGP models by prematurely removing these clusters.
In our experiments, we only merge clusters 
whose similarity exceeds a conservatively large merge threshold.
%

\subsection{Extension to EP}\label{sec:EPNRM}

For data sets of fixed size, $N$, ADF-NRM can be extended to \emph{EP-NRM} for batch
inference analogously to Sec. \ref{sec:EP}. 
Assume we have both an approximation to the
batch posterior $\hat{q}(\theta, z_{1:N})$ and local contributions
$\bar{q}_j(\theta,z_{j})$ for $j=1,\dots,N$, both of which can be computed
using ADF.  In particular,  $\hat{q}(\theta, z_{1:N}) = \hat{q}_N(\theta, z_{1:N})$, the final ADF posterior approximation, and $\bar{q}_j(\theta,z_{j}) \propto \frac{\hat{q}_{j}(\theta,z_{1:j})}{\hat{q}_{j-1}(\theta,z_{1:(j - 1)})}$, the ratio between successive ADF approximations.  Now define 
\begin{equation}\label{eq:local_rem}
    \hat{q}_{\backslash j}(\theta,z_{\setminus j}) \propto
    \frac{\hat{q}(\theta, z_{1:N})}{\bar{q}_j(\theta,z_j)}
\end{equation}
to be the approximate posterior with $x_j$ removed. We refine $\hat{q}(\theta,z_{1:N})$ using the two step approach outlined in Section \ref{sec:ADFNRM}, first appending the predictive factor, $p(z_j|z_{\setminus j})$, to  $\hat{q}_{\backslash j}(\theta,z_{\setminus j})$, followed by a projection step, and then adding the likelihood factor, $p(x_j|z_{j}, \theta)$, again followed by a projection step. Similar to ADF, the updated soft assignment for $z_j$ is given by $\hat{q}(z_{jk}) \propto q^{\pred}_{\setminus j} (z_{jk}) \int p(x_j| z_{jk}, \theta)
    \hat{q}_{\setminus j} (\theta_k) d\theta_k$ where $q^{\pred}_{\setminus j}$
    is the approximate predictive distribution given all other soft assignments.  
The global update is given by $\hat{q}(\theta_k) \propto p(x_j|z_{jk},\theta)^{\hat{q}(z_{jk})} \hat{q}_{\setminus j} (\theta_k)$.  The $j$th local contribution is
\begin{equation} \label{eq:local_update}
\bar{q}_j(\theta, z_ j) \propto
\frac{\hat{q}(\theta,z_{1:N})}{\hat{q}_{\backslash j}(\theta,z_{\backslash j})}.
\end{equation}
We cycle through the data set repeatedly, applying the steps above, until convergence. 

For conjugate exponential families, the computations required for the global cluster parameters, $\theta$, in Eq.~\eqref{eq:local_rem} and Eq.~\eqref{eq:local_update}  reduce to updating sufficient statistics as in Sec.~\ref{sec:EP}.  $q^{\pred}_{\setminus j}$ for NGGPs may similarly be updated on each round by letting
$S_{k}=\sum_{i = 1}^N \hat{q}(z_{ik})$ and $S_{k,\setminus j}=S_{k} - \hat{q}(z_{jk})$, where $\hat{q}(z_{ik})$ are the current soft assignments. 
Under the same logic as Eq. \eqref{eqn:nrmpredapprox}, $q^{\pred}_{\setminus j}$ for instantiated clusters is approximated by
\begin{align}
    q^{\pred}_{\backslash j} (z_{jk}) &\propto \max(S_{k, \setminus j} - \sigma,0),
\end{align}
and $q^{\pred}_{\backslash j} (z_{j,K+1})$ follows analogously (see Supplement). After computing the refined soft assignment, $\hat{q}(z_{jk})$, we update $S_k = S_{k, \setminus j} + \hat{q}(z_{jk})$. As a consequence of this approach, the total
weight on an instantiated cluster $k$,
$S_{k}$, can become
small upon revisits of the data
assignments. In practice, we remove cluster
$k$ if $S_k < \epsilon$, where $\epsilon$ is
as in Sec.~\ref{sec:coping}.

%% file: experiments.tex
\section{Experiments}\label{sec:experiments}

We evaluate ADF-NRM on both real and synthetic data using the task of document
clustering.  Each document is represented by a vector of word counts,
$x_d \in \reals_+^V$, where $V$ is the size of the vocabulary, and
$x_{dw}$ is the number of occurrences of word $w$ in document $d$. 
We then model the corpus as a NGGP mixture of multinomials; that is, our data are generated as in Eq.~\eqref{eqn:nrmmix} with $x_d \sim \MULT(N_d,\theta^{z_d})$, where $N_d$ is the number of words in document $d$ and $\theta_k$ is a vector of word probabilities in cluster $k$.  We take $H_0$ to be Dirichlet such that 
$\theta_k \sim \DIR(\alpha)$.  We then use
our proposed algorithms to perform inference over $\{z_d\}$ and
$\{\theta_k\}$.

We focus on comparing the IG ($\sigma=0.5$) to the DP ($\sigma=0$).  The choice of $\alpha$ for the Dirichlet base measure in our experiments are discussed in the Supplement.  To select the NRM hyperparameters $a$ and $\tau$, we adapt a grid-search method used for the batch sampling procedure of~\cite{Barrios:Lijoi:Nieto-Barajas:Prunster:2013} to our streaming setting.  As detailed in the Supplement, we perform a preliminary analysis on a small subset of the data.
Our algorithm is then let loose on the remaining data with these values fixed.


\subsection{Synthetic Bars}
\begin{figure}[t]
        \centering
        \begin{subfigure}[b]{0.23\textwidth}
                \includegraphics[width=\textwidth]{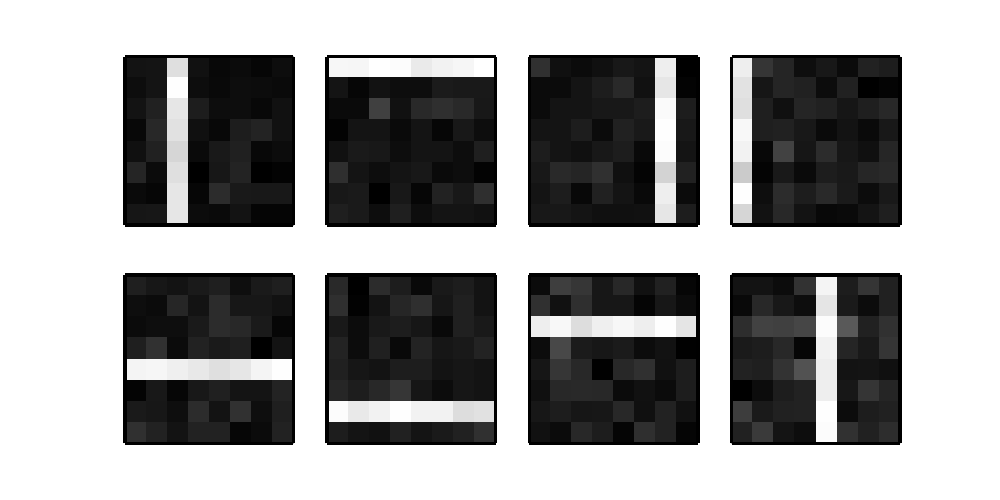}
        \end{subfigure} \\
        \vspace{-10 pt}
        \begin{subfigure}[b]{0.45\textwidth}
                \includegraphics[width=\textwidth]{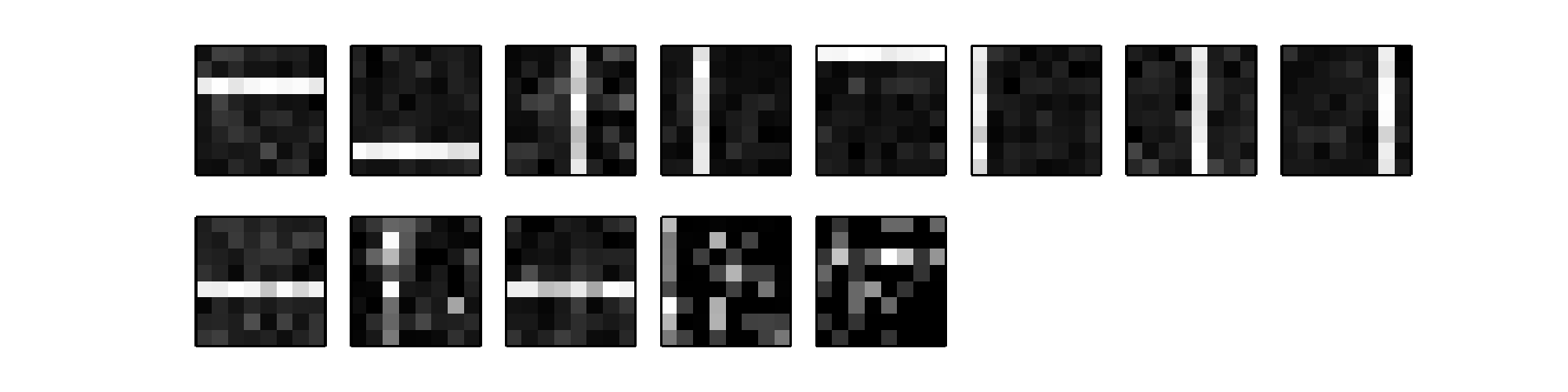}
        \end{subfigure}
\precap
\vspace{10 pt}
        \caption{\small ADF-NRM posterior mean mixture components for the bars data with (\emph{top}) and without (\emph{bottom}) merge.} \postcap
        	\label{figure:bars}
\end{figure}
First, we perform clustering on a synthetic data set of $8 \times 8$
images to show that ADF-NRM can recover the correct
component distributions.  Each image is represented by a vector of positive integer pixel intensities,
which we interpret as a document over a vocabulary with $64$
terms.  The clusters correspond
to horizontal and vertical bars with an additive
baseline to ensure cluster overlap.
Each of 200 images is generated by first choosing a cluster, $z_d$,
and then sampling pixel intensities $x_d \sim \MULT(50, \theta^{z_d})$.
Fig.~\ref{figure:bars} depicts the resulting ADF-NRM posterior mean mixture components under
the learned variational distribution, $\E_{\hat{q}_N}[\theta_k]$, based on an IG prior ($\sigma=0.5$), both with and without merge
moves.  We
see that in both cases the algorithm learns the correct clusters, but merge
moves remove redundant and extraneous clusters.

\subsection{Synthetic Power-Law Clusters}
\label{sec:synth_pl}

To explore the benefit of the additional flexibility of IGs over DPs, we generated 10,000 synthetic
documents, $x_{d}$, from a Pitman-Yor$(.75,1)$ mixture of multinomials.  The Pitman-Yor prior is
another commonly used BNP prior famous for its ability to model
clusters whose sizes follow certain power-law distributions~\cite{Pitman:1997}.

We assess the ADF-NRM predictive log-likelihood and inferred number of clusters versus number of observed
documents. For each model, we selected hyperparameters based on a randomly selected set of $1,000$ documents. We then continue our algorithm on $7,000$ training documents and use the remaining $2,000$ for evaluation. Mean predictive log-likelihoods, number of clusters, and error estimates were obtained by permuting the order of the training documents 5 times.  We compare our ADF-NRM performance to that of a baseline model where the cluster parameters are inferred based on ground-truth-labeled training data. Lastly, after the completion of ADF, we performed 49 additional passes through the data using EP-NRM to obtain refined predictions and number of clusters.

\begin{figure}[t]
        \centering
        \begin{subfigure}[b]{0.24\textwidth}
                \includegraphics[width=\textwidth]{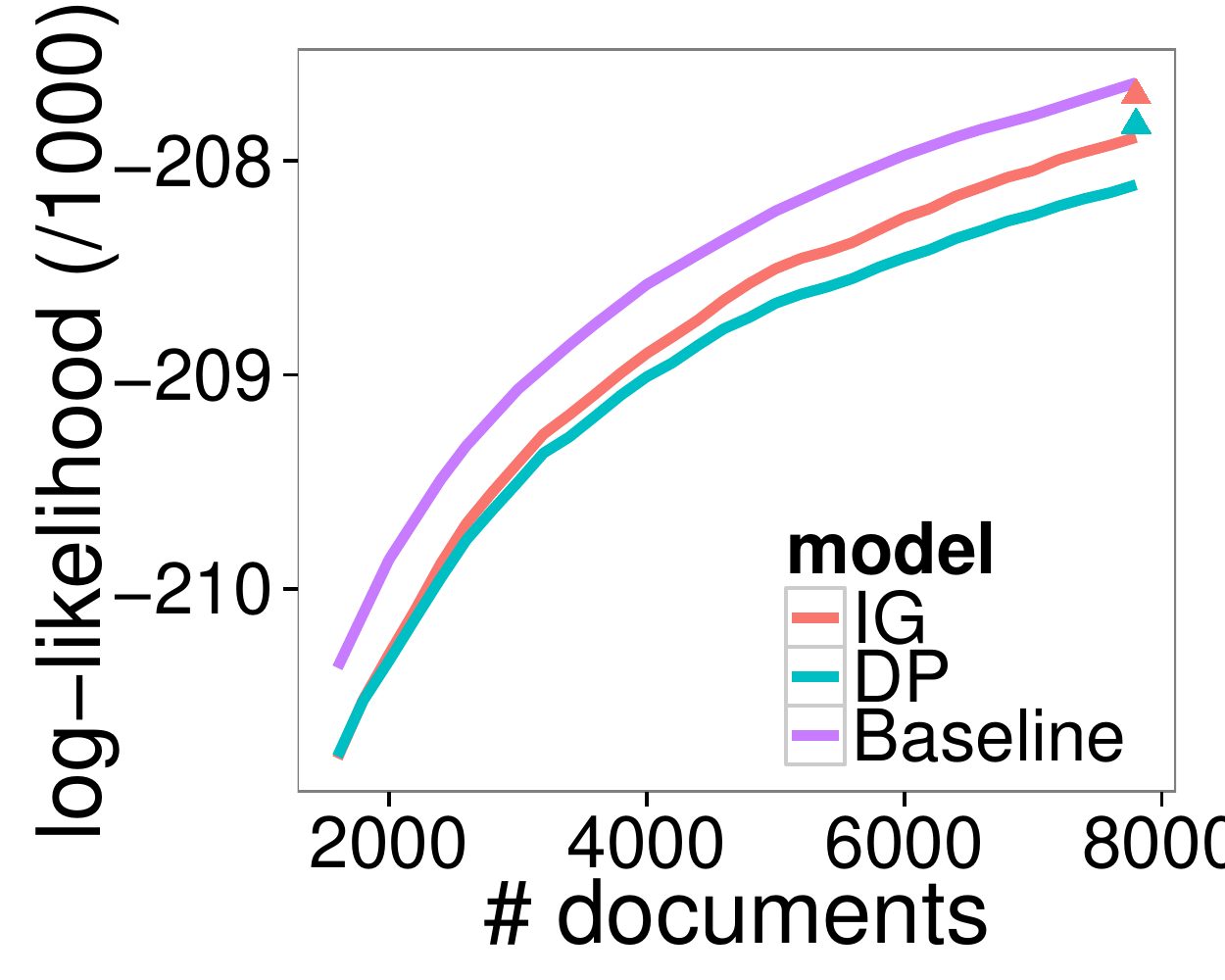}
        \end{subfigure}%
        \begin{subfigure}[b]{0.24\textwidth}
                \includegraphics[width=\textwidth]{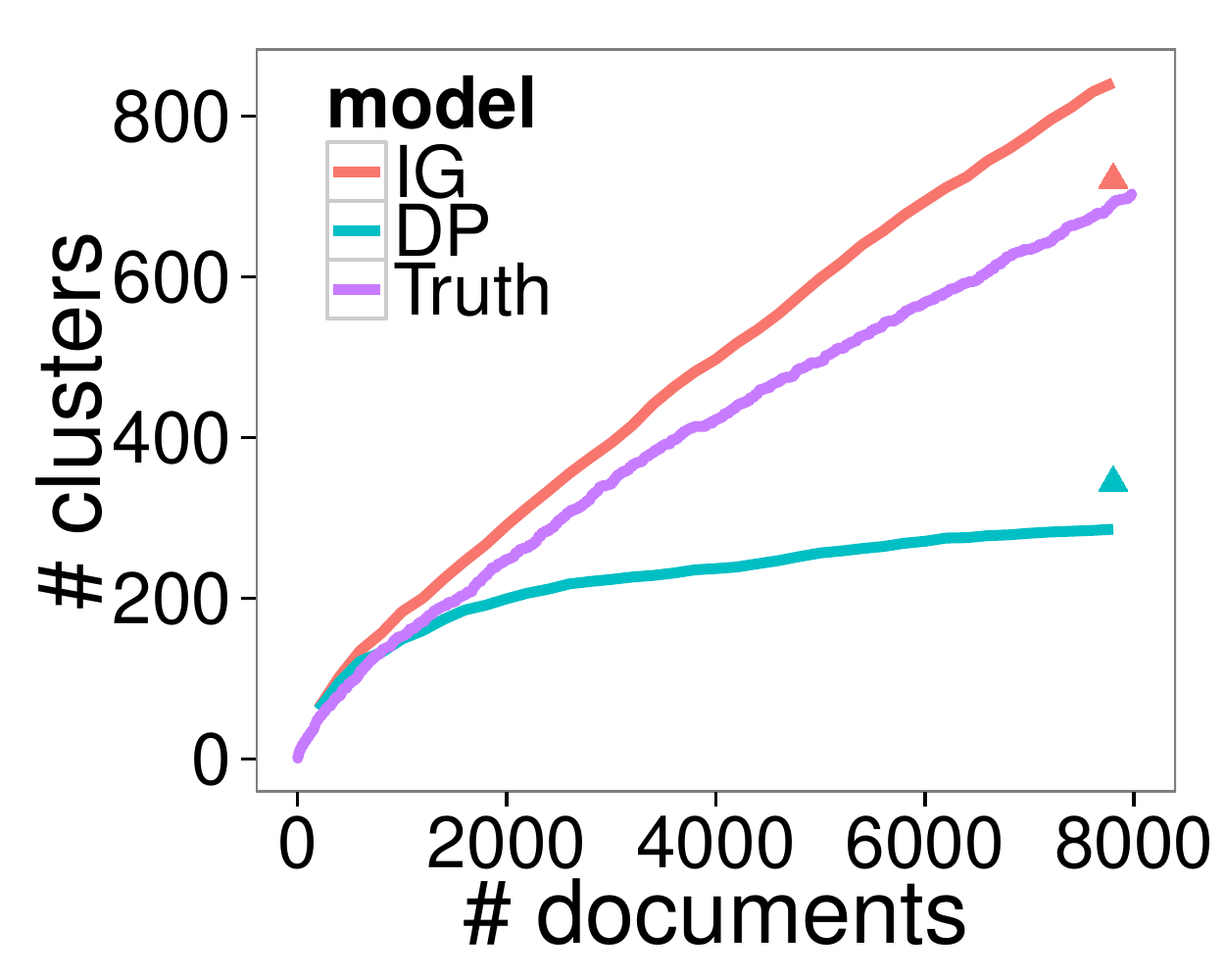}
        \end{subfigure}
\precap
        \caption{\small Mean predictive log-likelihood (\emph{left}) and number
        of clusters (\emph{right}) for the DP (\emph{cyan}) and IG (\emph{red}) priors on the synthetic power-law
        data set using ADF-NRM. Triangles indicate final values for
    EP-NRM after 50 epochs. The ground-truth model is shown in purple. Error bars are omitted due to their small size relative to the plot scale.}\postcap     
        	\label{fig:synth_kos}
\end{figure}

We see in Fig.~\ref{fig:synth_kos}
that both the IG and DP models perform similarly for small $n$, but as the amount of data
increases, the IG provides an increasingly better fit in terms of both predictive
log-likelihood and number of clusters. This substantiates the importance of our streaming algorithm being able to handle a broad class of NRMs.  Furthermore, after a single data pass, ADF-NRM comes close to reaching the baseline model even with the IG/Pitman-Yor model mismatch. It is also evident in Fig.~\ref{fig:synth_kos} that additional EP iterations both improve predictions and the match between inferred and true number of clusters for both prior specifications. 

\subsection{KOS Blog Corpus}
\begin{table}[t] 
\label{tab:kos}
\caption{\small Mean predictive performance and number of
clusters ($\pm$ 1 std. err.) for ADF-NRM, EP-NRM, and a collapsed Gibbs sampler on the KOS
corpus.}
\vspace{-0.1in}
\label{tab:kos}
\begin{center}
\begin{tabular}{|c|c|c|c|} \hline
\small
{ \bf Method} & {\bf Pred. log-lik} & {\bf \#Clusters} & {\bf Epochs} \\
\hline\hline
ADF-DP & -346023 $\pm$ 165 & 80  $\pm$ .17 & 1 \\ \hline
ADF-IG & -345588  $\pm$  159 & 92 $\pm$ .18 & 1 \\ \hline
EP-DP & -342535 $\pm$ 181 & 104 $\pm$ 2.4 & 50 \\  \hline
EP-IG & -342195 $\pm$ 161 & 114 $\pm$ 1.5 & 50  \\ \hline
Gibbs-DP & -342164 $\pm$ 11 & 119 $\pm$ 0.3 & 215 \\ \hline
Gibbs-IG & -341468 $\pm$ 338 & 128 $\pm$ 1.3 & 215 \\ \hline
\end{tabular}
\end{center}
\label{default}
\vspace{-0.15in}
\end{table}

We also applied ADF-NRM to cluster the KOS
corpus of 3,430 blog posts~\cite{Bache:Lichman:2013}.  The fact that the corpus is small
enough to use non-streaming (batch) inference algorithms allows us to compare
ADF-NRM, EP-NRM, and the collapsed Gibbs sampler
for NGGP mixture models presented in~\cite{Favaro:Teh:2013}. Importantly, we only compare to Gibbs, which is not suited to the streaming setting, in an attempt to form a gold standard. (Recall that Gibbs targets the exact posterior
in contrast to our variational-based approach, and we do not expect mixing to be an issue in this modest-sized data set.)


We evaluated performance as in
Sec.~\ref{sec:synth_pl}.  Here, we held out $20\%$ of the
entire corpus as a test set and trained (given the hyperparameters determined
via grid search) on the remaining $80\%$ of documents.  The ADF-NRM predictive log-likelihoods for the IG and DP were computed after a single pass through the data set while
those for EP-NRM were computed by cycling
through the data set 50 times. Error estimates were obtained by permuting the order of the documents 20 times. Predictions for the collapsed Gibbs sampler were computed by running 5 chains for 215 passes
through the data and averaging the predictive log-likelihood for the last 50
samples across chains. 

\begin{figure}[t]
	\centering
	\begin{subfigure}[b]{0.25\textwidth}
                \includegraphics[width=\textwidth]{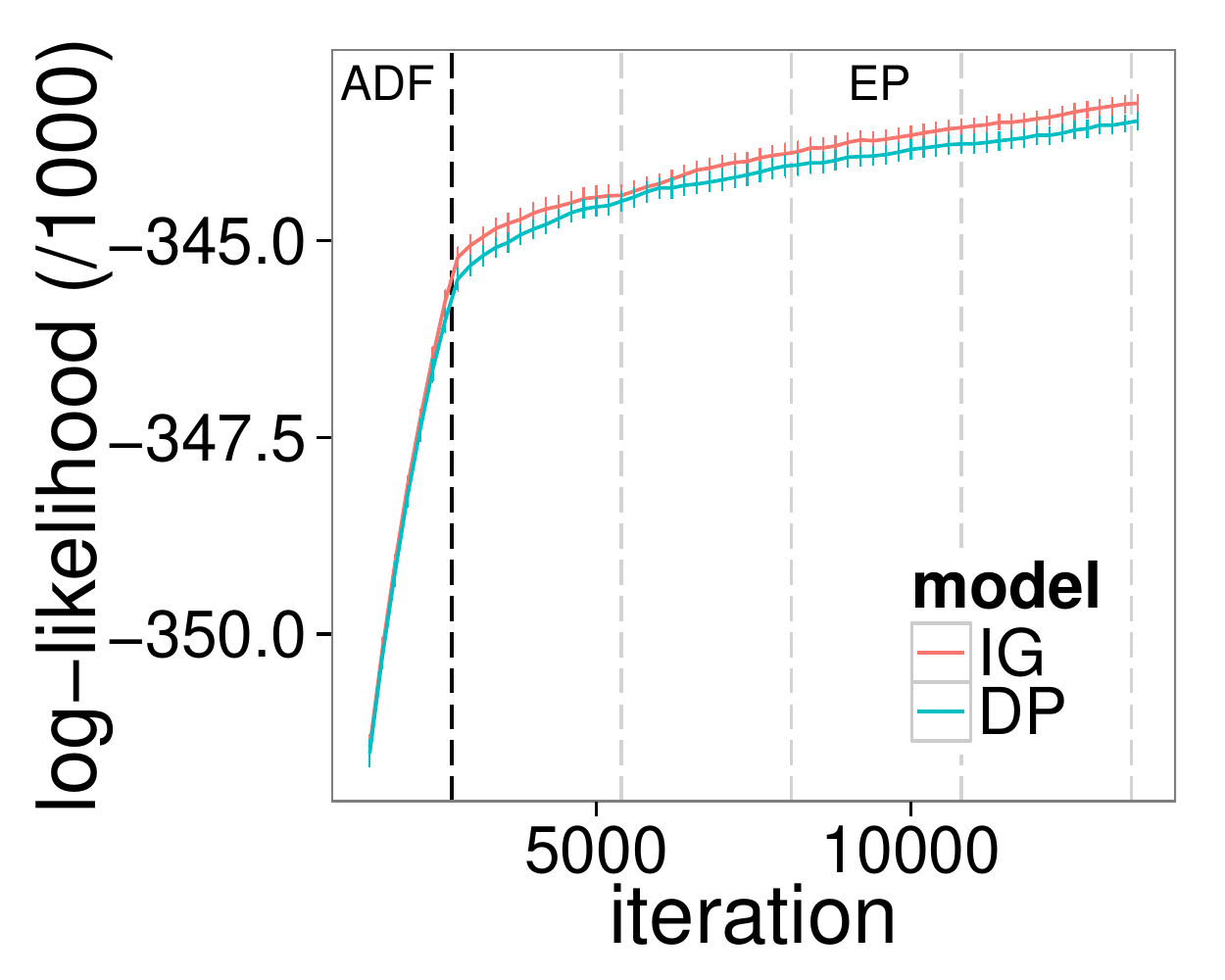}
        \end{subfigure}%
        \begin{subfigure}[b]{0.25\textwidth}
                \includegraphics[width=\textwidth]{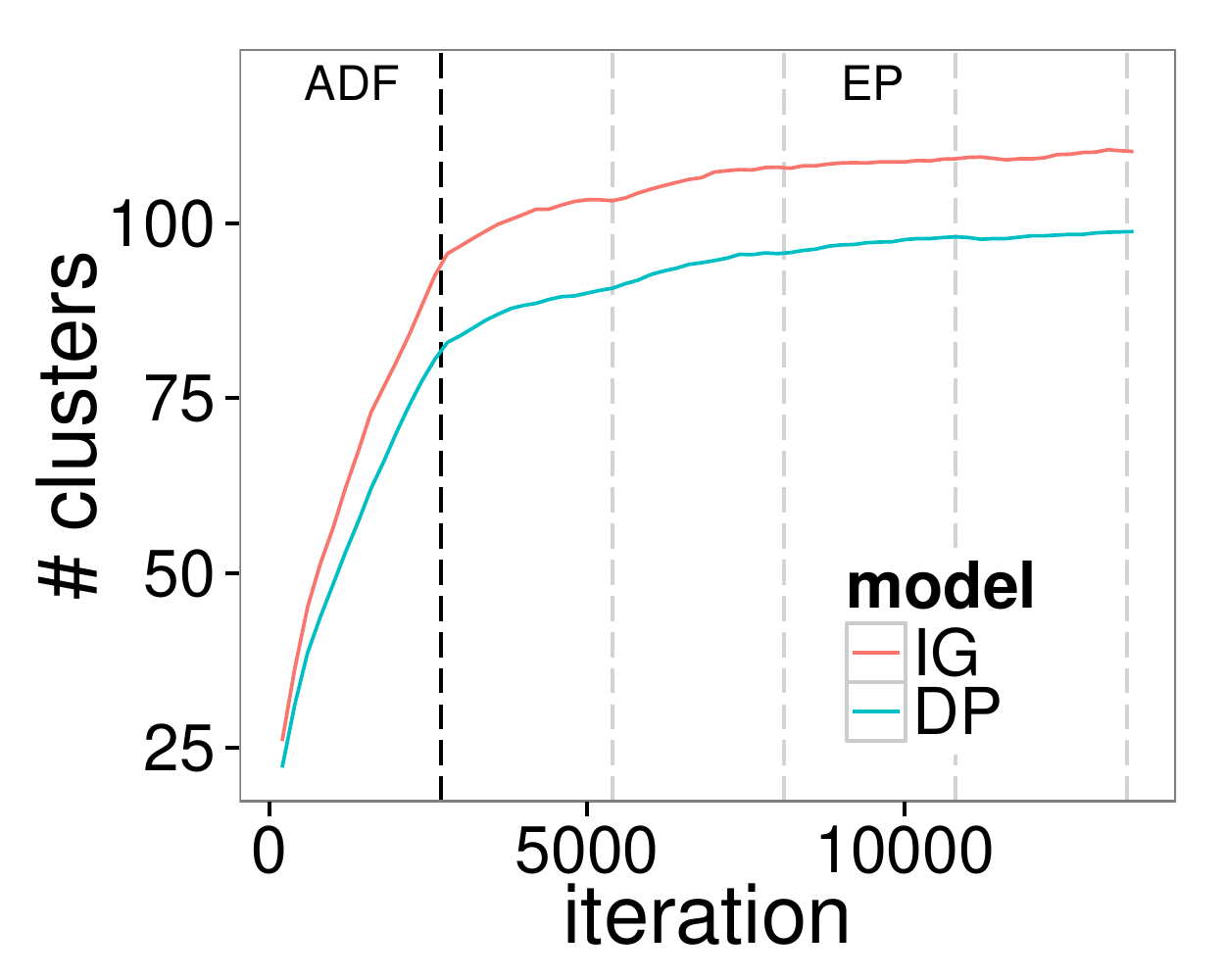}
        \end{subfigure}
\precap
        \caption{\small Predictive log-likelihood (\emph{left}) and mean number
        of clusters (\emph{right}) using EP-NRM on KOS corpus. Vertical lines
indicate epochs and error bars $\pm$ 1 st. dev.. }\postcap 
        \label{fig:EP}
\end{figure}

The comparisons between all methods are depicted in Table~\ref{tab:kos}.  For
all algorithms (ADF, EP, and Gibbs) the added flexibility of the IG provides a
better fit in terms of predictive log-likelihood.  The extra $\approx 10$
clusters associated with the IG for all algorithms correspond to small clusters 
which seem to capture finer-scale latent structure important for prediction.  Although
performance increases moving from the one-pass ADF-NRM to multi-pass EP-NRM,
Fig.~\ref{fig:EP} shows that the most gains occur in the first
epoch.  In fact,  after one epoch ADF performs significantly
better than a single epoch of Gibbs; it takes about three Gibbs epochs to reach comparative performance (see Supplement). 
Finally, while the IG Gibbs sampler leads to the
best performance, EP-NRM with the IG prior is competitive and reaches similar performance to
the DP using Gibbs.

In summary, ADF-NRM provides
competitive performance with only a single pass
through the data; refined approximations nearly matching the
computationally intensive samplers can be computed via EP-NRM
if it is feasible to save and cycle through the data.

\vspace{-.1in}

\subsection{New York Times Corpus}
\vspace{-.1in}
We performed streaming inference on a corpus of 300,000 New York Times
articles~\cite{Bache:Lichman:2013}.  We first identified a
vocabulary of $7,841$ unique words by removing words occurring in fewer than 20
and more than $90\%$ of documents, as well as terms resulting from
obvious errors in data acquisition.  Then, we removed documents containing
fewer than 20 words in our vocabulary, resulting in a corpus of 266,000
documents.  
The corpus is too large for batch algorithms, so
we focus on ADF-NRM comparing the DP and IG priors.

\begin{figure}[t]
	\centering
	\begin{subfigure}[b]{0.25\textwidth}
                \includegraphics[width=\textwidth]{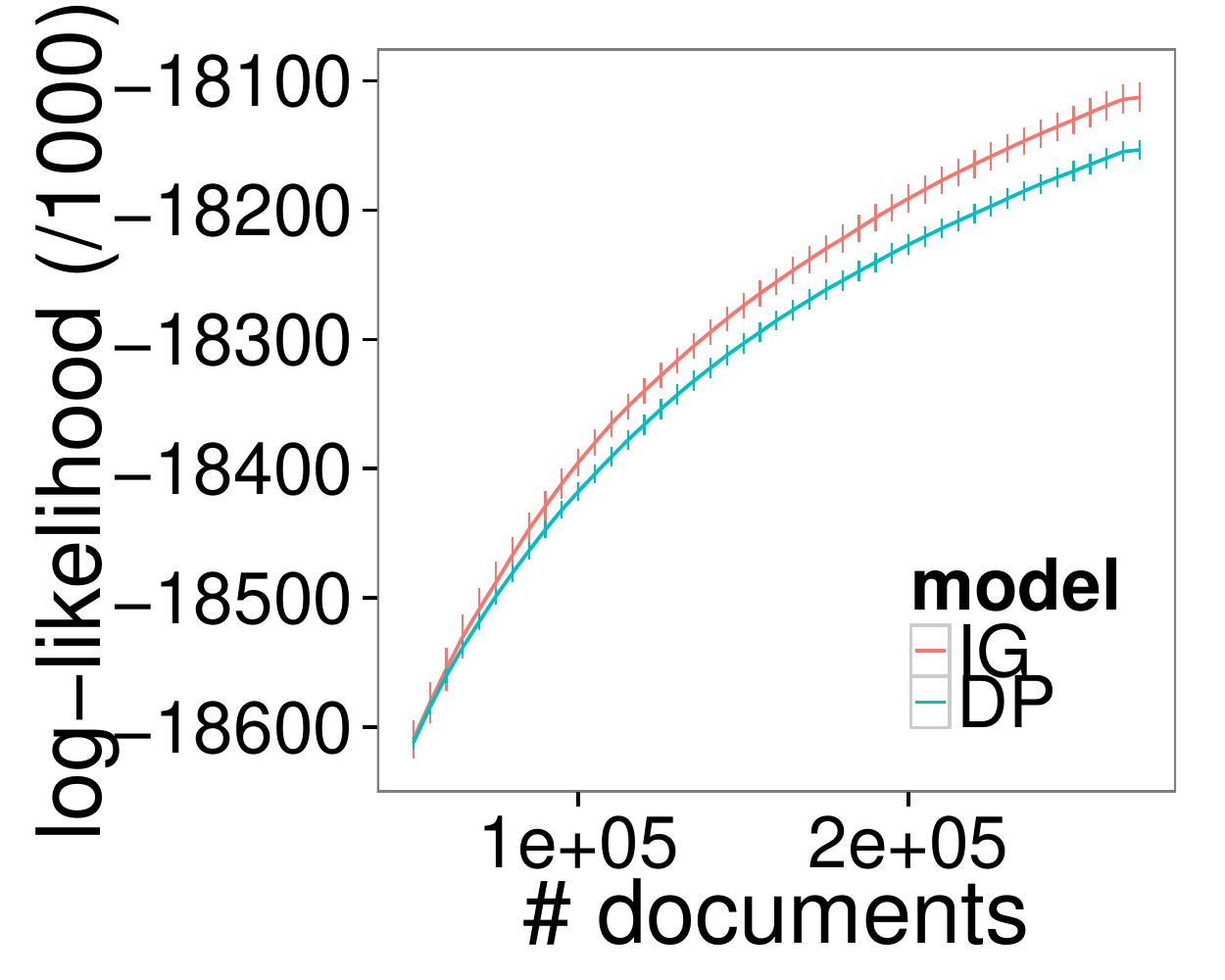}
        \end{subfigure}%
        	\begin{subfigure}[b]{0.25\textwidth}
                \includegraphics[width=\textwidth]{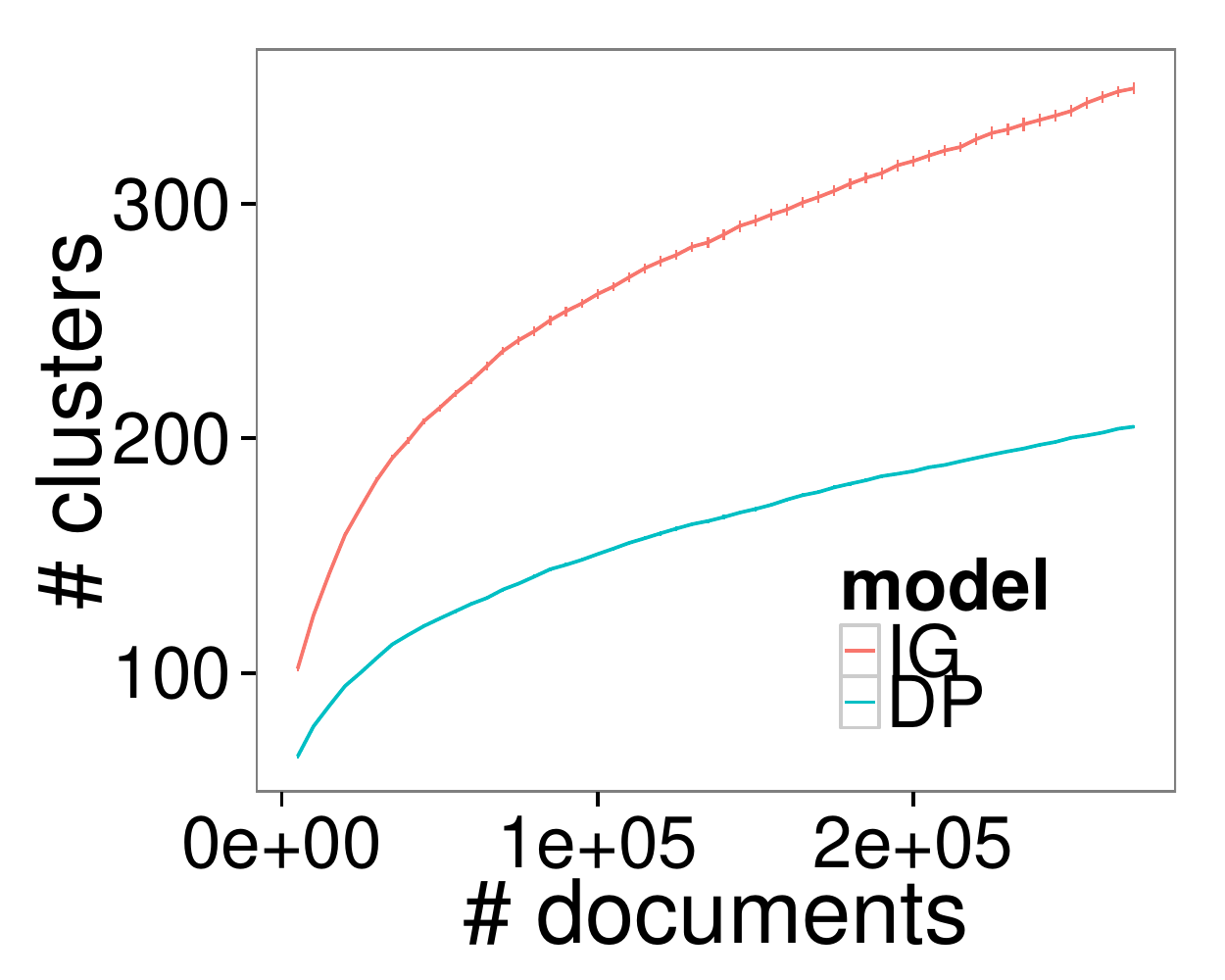}
        \end{subfigure}
\precap
        \caption{\small Comparison of (\textit{left}) predictive log-likelihood and (\textit{right}) 
            number of clusters using ADF-NRM on the New York Times corpus
        for the IG and DP priors.}\postcap
        \label{fig:nyt}
\end{figure}

\begin{figure}[t]
	\centering
	
                \includegraphics[width=\columnwidth]{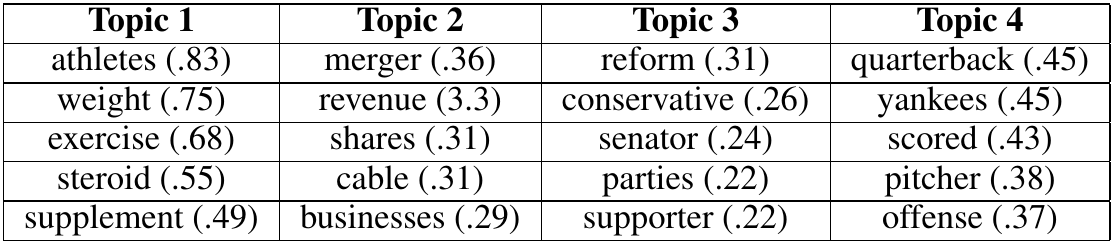}
\precap
        \caption{\small Most probable words and their respective contributions (in  \%) for the $4$
				         most prevalent topics.}\postcap
        \label{fig:nyt_topics}
\end{figure}

We determined hyperparameters as before and held out $5,000$ documents as a
test set, evaluating the predictive log-likelihood and number of clusters after
every $5,000$ training documents were processed.
See Fig.~\ref{fig:nyt}. As before, the IG obtains
superior predictive log-likelihood and introduces many additional small clusters compared to the DP, suggesting that the IG may be capturing
nuanced latent structure in the corpus that the DP cannot (see the Supplement for details).  Reassuringly, the
recovered clusters with highest weights correspond to interpretable topics (Fig.~\ref{fig:nyt_topics}).
Again, we
see the benefits of considering NRMs beyond the DP, which has been the most used BNP prior due to the
computational tools developed for it.

%% file: discussion.tex
\section{Discussion}

We introduced the ADF-NRM algorithm, a variational approach to streaming approximate posterior inference in NRM-based
mixture models. Our algorithm leverages the
efficient sequential updates of ADF while importantly maintaining
the infinite-dimensional nature of the BNP model. The key to tractability is focusing on approximating a partial-urn characterization of the NRM predictive distribution of cluster
assignments.  We also showed how
to adapt the single-pass ADF-NRM algorithm to a multiple-pass EP-NRM variant
for batch inference.  Our empirical results demonstrated the effectiveness of our algorithms, and the importance of considering NRMs beyond the DP.

A potential drawback of the EP-NRM scheme is that each observation needs to store its
variational distribution over cluster assignments. 
An interesting question is if the local distributions
can be grouped and memoized~\cite{Hughes:Sudderth:2013} to save
computation and perform data-driven split-merge moves. This combined with 
parallel EP \cite{Gelman:2014, Xu:2014} would scale EP-NRM to massive data sets.  


Instead of using predictive distributions and exploiting the NRM partial-urn scheme, a natural question is if similar algorithms can be
developed that do not integrate out the underlying measure. Such algorithms
would be applicable to hierarchical BNP models such as topic models
and hidden Markov models~\cite{Teh:Jor:2010a}.

{\small
{\bf Acknowledgements:} This work was supported in part by DARPA Grant FA9550-12-1-0406 negotiated by AFOSR, ONR Grant N00014-10-1-0746, and the TerraSwarm Research Center sponsored by MARCO and DARPA. AT was partially funded by an IGERT fellowship.
}

%% file: supplementADF.tex
\section{Derivation of Global Update}
This section motivates the use of the mean field update for the global variables,  given in Eq. (18) of the main text, as an approximation to the optimal update for ADF after adding in the likelihood factor.  The presentation adapts that of \cite{Wang:Blei:2012} for ADF-NRM. 

Let $\hat{p}(\theta, z_{1:n} | x_{1:n}) \propto p(x_n|\theta, z_n) q^{\text{pr}}(z_{1:n}, \theta)$ denote the approximate posterior under the past variational updates after adding in the $n$th observation/likelihood factor.  The optimal $q(\theta_k)$ under ADF is given by the marginal distribution of $\hat{p}$:
\begin{align}
\hat{q}_n(\theta_k) \propto & \int \sum_{z_{1:n}}  \hat{p}(\theta_k| \theta_{\setminus k}, z_{1:n}, x_{1:n}) \times \\
&\hat{p}(\theta_{\setminus k}|z_{1:n}, x_{1:n}) \hat{p}(z_{1:n}|x_{1:n}) d \theta_{\setminus k}. \nonumber
\end{align}
Both the sums and integrals are intractable so we use the approximations: $\hat{p}(\theta_{\setminus k}|z_{1:n}, x_{1:n}) \approx \hat{q}_n(\theta_{\setminus k})$ and $\hat{p}(z_{1:n}|x_{1:n}) \approx \hat{q}_n(z_{1:n}) = \prod_{i = 1}^n \hat{q}_n(z_{i})$ which yields:
\begin{align}
\hat{q}_n(\theta_k) & \stackrel{\approx}{\propto} \int \sum_{z_{1:n}}  \hat{p}(\theta_k| \theta_{\setminus k}, z_{1:n}, x_{1:n}) \hat{q}_n(\theta_{\setminus k}) \hat{q}_n(z_{1:n}) d \theta_{\setminus k} \nonumber \\
&= E_{\hat{q}_n(z_{1:n}), \hat{q}_n(\theta_{\setminus k})} [\hat{p}(\theta_k| \theta_{\setminus k}, z_{1:n}, x_{1:n})] \nonumber \\
&= \exp \{ \log E_{\hat{q}_{n}(z_{1:n}), \hat{q}_{n}(\theta_{\setminus k})} [\hat{p}(\theta_k| \theta_{\setminus k}, z_{1:n}, x_{1:n})] \} \nonumber \\
&\leq \exp \{ E_{\hat{q}_{n}(z_{1:n}), \hat{q}_{n}(\theta_{\setminus k})} [\log  \hat{p}(\theta_k| \theta_{\setminus k}, z_{1:n}, x_{1:n})] \} \nonumber \\
&\propto \exp \{ E_{\hat{q}_{n}(z_{1:n}), \hat{q}_{n}(\theta_{\setminus k})} [\log  \hat{p}(\theta, z_{1:n}|x_{1:n})] \} \label{eqn::meanfield}
\end{align}

where the inequality follows by Jensen's inequality \cite{Wang:Blei:2012}. The approximation is tight when $\hat{q}(z_{1:n})$ and $\hat{q}(\theta_{\setminus k})$ approach Dirac measures.  Eq.~\eqref{eqn::meanfield} is that of the standard mean field update for $\hat{q}(\theta_k)$ \cite{Bishop:2006}. Since the $q(\theta_k)$ distributions are unknown for all $k$, we could perform coordinate ascent and cycle through these updates for each of the $\theta_k$ given the other $\theta_{\setminus k}$ and $\hat{q}(z_{1:n})$. Conveniently, since the $\hat{q}(z_{1:n})$ is already optimized by its tractable marginal, the $\theta_k$s are conditionally independent given the assignments in the mixture model, and  $q^{\text{pr}}(z_{1:n}, \theta) = q^{\text{pr}}(z_n) \prod_{i = 1}^{n - 1} \hat{q}_{i}(z_{i}) \prod_{k = 1}^{\infty} \hat{q}_{n-1}(\theta_k)$, we can perform a single mean field update for each $\theta_k$ given by,
\begin{equation}
\hat{q}_n(\theta_k) \propto p(x_n |z_{nk}, \theta)^{\hat{q}_n(z_{nk})} \hat{q}_{n-1}(\theta_k).
\end{equation}

%% file: nrm_pred_general.tex
\section{Derivation of approximate NRM predictive rule}

In this section we provide the derivation of $q^{\pred}(z_n)$ for NRMs given in
Eq. (25) of the main text.  We start by presenting the derivation for general
NRMs and then demonstrate how to apply ideas to NGGPs.  The presentation in
this section is adapted from~\cite{Favaro:Teh:2013,James:Lijoi:Prunster:2009}.

\subsection{General NRMs}

We assume the mixture model specification in Eq.~(3) from the main text.  In
particular we note that the unnormalized mixture weights $\pi = (\pi_1, \pi_2,
\ldots)$ are drawn from a completely random measure with \Levy measure
$\lambda(d\pi)$.  We also introduce the \textit{exponentially tilted \Levy
measure} as $e^{-U\pi}\lambda(d\pi)$ which will appear below.

First, we expand the sum in the approximate
predictive distribution, $q^{\mathrm{pr}}(z_n)$, to include the unnormalized masses, $\pi$, and the
auxiliary variable $U_{n-1}$:
\begin{align}
q^{\pred}(z_n) &= \sum_{z_{1:n-1}} p(z_n|z_{1:n-1}) \prod_{i = 1}^{n-1} \hat{q}_{n-1}(z_i) \\
&= \sum_{z_{1:n-1}} \iint p(z_n|\pi) p(\pi|U_{n-1}, z_{1:n-1}) \\
& \times p(U_{n-1}|z_{1:n-1}) dU_{n-1} d\pi \prod_{i = 1}^{n-1} \hat{q}_{n-1}(z_i) \label{eqn:nrmpred} \nonumber
\end{align}
where the conditional distribution of the auxiliary variables $U_{n-1}$ given
the past assignments is given
by:
\begin{equation} \label{eq:Ucond}
p(U_{n-1}|z_{1:n-1}) = U_{n-1}^{n-1} e^{-\phi(U_{n-1})} \prod_{k = 1}^{K_{n-1}} \kappa_{n_k}(U_{n-1})
\end{equation}
where $\phi(U)$ is the Laplace exponent of the underlying CRM,
$\phi(U) = \int (1 - e^{-Us}) \lambda(ds)$, and $\kappa_{m}(U)$ denotes the
$m$th moment of the exponentially tilted \Levy measure,
$\kappa_{m} = \int s^{m} e^{-Us} \lambda(d s)$.

Let $K_{n - 1}$ denote the number of components considered for the observations
$z_{1:n-1}$.
The conditional distribution of the random measure,
$\pi = (\pi^*, \pi_1, \ldots, \pi_{K_{n-1}})$,  given $U_{n-1}$ and the
assignments, $z_{1:n-1}$, is:
\begin{align}
p(\pi| U_{n-1}, z_{1:n-1}) &= p(\pi^*|U_{n-1}) \prod_{k = 1}^{K_{n-1}}
p(\pi_k|z_{1:n-1}, U_{n-1}).
\end{align}
where $\pi_{1:K_{n-1}}$ are the masses of all the instantiated components
and $\pi^*$ denotes the mass assigned to the uninstantiated components.
The distribution of $\pi_k$ is given by
\begin{align}
p(\pi_k|z_{1:n-1}, U_{n-1}) \propto \pi_k^{n_k} e^{-U_{n-1} \pi_k} \lambda(d \pi_k),
\end{align}
where $n_k$ is the number of observations assigned to component $k$ in
$z_{1:n-1}$ and $\pi^*$ follows a Poisson process (PP) with exponentially
tilted \Levy measure, $\pi^* \sim \POISP(e^{-U_{n-1}\pi^*} \lambda(d\pi^*))$,
where again $\lambda(ds)$ is the \Levy measure of the unnormalized masses.
Since the integral in
Eq~\eqref{eqn:nrmpred} is intractable, we introduce a variational approximation
for $\pi$ and $U_{n-1}$. In particular,
we use a partially factorized approximation
\begin{align} \label{eqn:uvar}
p(\pi|U_{n-1}, z_{1:n-1}) p(U_{n-1}|z_{1:n-1}) \hat{q}(z_{1:n-1}) \approx \\
q(\pi|U_{n-1}) q(U_{n-1}) \hat{q}(z_{1:n-1}), \nonumber
\end{align}
where $\hat{q}(z_{1:n-1}) = \prod_{i=1}^{n-1}\hat{q}_{n-1}(z_i)$ is fixed and
given from previous iterations.  We
perform a mean field step to minimize the KL divergence between the left and right
hand sides of Eq \eqref{eqn:uvar}.   Specifically, we compute the optimal
$q(U_{n-1})$ and then given that we compute the optimal $q(\pi|U_{n-1})$.
Because of the factorization given in the left hand of Eq.~\eqref{eqn:uvar}
this procedure gives the optimal distributions. According to standard mean
field updates the optimal distribution for $q(U_{n-1})$ is given by:
\begin{equation}
    \log q(U_{n-1}) = E_{\hat{q}(z_{1:n-1})} \log p(U_{n-1}|z_{1:n-1}) + C
\end{equation}
where $p(U_{n-1}|z_{1:n-1})$ is given in Eq. \eqref{eq:Ucond}. The tractability of this variational approximation for $U_{n-1}$ 
will depend on the NRM under consideration.  For the NGGP it is conveniently
given in closed form, as detailed below in Section \ref{Sec:forNGGP}.  However,
efficient numerical algorithms can be used to compute the necessary integrals
for general NRMs.

Given the optimal $q(U_{n-1})$, the optimal variational approximations to the
masses,
$q(\pi|U_{n-1}) =  q(\pi^*|U_{n-1}) \prod_{i=1}^{K_{n-1}} q(\pi_j|U_{n-1})$,
are given by
\begin{equation} \label{eqn:pik}
q(\pi_k|U_{n-1}) \propto \pi_k^{\E_{\hat{q}}[n_k]} e^{-U \pi_k} \lambda (d \pi_k)  \mbox{      for $k = 1 \ldots K_{n-1}$},
\end{equation}
where $\E_{\hat{q}}[n_k]$ is the expected number of assignments to component $k$ and is given by:
\begin{align}
\E_{\hat{q}}[n_k] = \sum_{i = 1}^{n-1} \hat{q}(z_{ik}).
\end{align}
Under $q(\pi^*|U_{n-1})$, $\pi^*$ is still drawn from $\POISP(e^{-U_{n-1}w}
\lambda(dw))$. Using these approximations Eq.~\eqref{eqn:nrmpred} becomes
\begin{align}
q^{\pred}(z_n) &= \sum_{z_{1:n-1}} \iint p(z_n|\pi) p(\pi|U, z_{1:n-1}) \\
& \times p(U_{n-1}|z_{1:n-1}) dU_{n-1} d\pi \prod_{i = 1}^{n-1} \hat{q}_{n-1}(z_i) \nonumber \\
&\approx  \iint p(z_n|\pi) q(\pi|U_{n-1}) q(U_{n-1}) d\pi dU_{n-1} \label{eqn::medint} \\
&= \int  q(z_n|U_{n-1}) q(U_{n-1}) dU_{n-1} \label{eqn::finalint}
\end{align}
where 
\begin{equation} \label{eq:prednrm}
q(z_{nk}|U_{n-1}) \propto
\begin{cases}
\max \left(\frac{\kappa_{E_{\hat{q}}[n_k] + 1}(U_{n-1})}{\kappa_{\E_{\hat{q}}[n_k]}(U_{n-1})} ,0 \right),  \, \text{$k \leq K_{n-1}$} \\
\kappa_{1}(U_{n-1}),  \,\,\,\,\,\,\,\,\,\,\,\,\,\,\,\,\,\,\,\,\,\,\,\,\,\,\, \text{$k = K_{n-1} + 1$}.
\end{cases}
\end{equation}
Eq. (\ref{eqn::finalint}) arises from (\ref{eqn::medint})
by an application of Prop. 2.1 in \cite{Favaro:Teh:2013}.  In
Eq.~\eqref{eq:prednrm}, the maximum with
zero is necessary since if the expected number of
clusters assigned to a cluster $k$, $\E_{\hat{q}}[n_k]$, is small then the
variational distribution for $\pi_k$ given in
Eq.~\eqref{eqn:pik} might be degenerate at zero and so there will be zero
probability of a new observation being assigned to that cluster.  More details
for the NGGP case are given in Section~\ref{Sec:forNGGP}.

\subsection{Predictive Rule for the NGGP}
 \label{Sec:forNGGP}

For NGGPs, the general equations for NRMs described above reduce to simple,
analytically tractable forms.  In particular, the variational approximation
$q(U_{n-1})$ is given by
\begin{align}
q(U_{n-1}) \propto \frac{U_{n-1}^{n-1}}{(U_{n-1} + \tau)^{n - 1 - a \E_{\hat{q}(z_{1:n-1})}[K_{n-1}']}} e^{-\frac{a}{\sigma}(U_{n-1} + \tau)^{\sigma}}
\end{align}
where $\E_{\hat{q}(z_{1:n-1})}[K'_{n-1}]$ is the expected number of clusters
instantiated thus far. This expectation is given by:
\begin{align} \label{eq:Eclust}
\E_{\hat{q}(z_{1:n-1})}[K_{n-1}'] &= K_{n-1} - \sum_{j=1}^{K_{n-1}} \left(\prod_{i = 1}^{n - 1} (1 - \hat{q}(z_{ij})) \right) \\
&\stackrel{n \to \infty}{\to} K_{n-1}.
\end{align}
Note that Eq. \eqref{eq:Eclust} does not require all past soft assignments to be
saved; instead, only $\prod_{i = 1}^{n - 1} (1 - \hat{q}(z_{ij}))$
must be stored for each component and updated after each observation. In
practice we find that using
$\E_{\hat{q}(z_{1:n-1})}[K_{n-1}'] \approx K_{n-1}$ leads to comparable
performance to evaluating the complete expectation. This occurs because, given
our thresholding scheme for mixture components, each component has a few
$\hat{q}(z_{ik})$ that are close to one, making the product close to zero. 

Additionally, in the case of the NGGP the $\kappa_m(U)$ functions needed in
Eq.~\eqref{eqn::finalint} are given by
\begin{equation}
    \kappa_m(U) = \frac{a}{(U + \tau)^{m-\sigma}}
        \frac{\Gamma(m-\sigma)}{\Gamma(1-\sigma)},
\end{equation}
which when plugged into Eq.~\eqref{eq:prednrm} yields
\begin{equation}
q(z_{nk}|U_{n-1}) \propto
\begin{cases}
\max \left( \sum_{i = 1}^{n-1} \hat{q}(z_{ik}) - \sigma,0 \right),  \, \text{$k \leq K_{n-1}$} \\
a(U_{n-1} + \tau)^{\sigma},\,\,\,\,\,\,\,\,\,\,\,\,\,\,\,\,\,\,\,\,\,\,\text{$k = K_{n-1} + 1$}.
\end{cases}
\end{equation}
When we approximate the integral in Eq.~\eqref{eqn::finalint} with a delta
function about the maximum, $\hat{U}_{n-1} = \arg \max q(U_{n-1})$ we see that
$q^{\mathrm{pr}}(z_{nk}) \approx q(z_{nk}|\hat{U}_{n-1})$, which is exactly
Eq.~(25) of the main text.  Alternatively, one could compute the
integral in Eq.~\eqref{eqn::finalint} numerically by first performing a change
of variables, $V_{n-1} = \log U_{n-1}$, to obtain a log-convex density over
$V_{n-1}$ and then use adaptive rejection sampling to sample from $V_{n-1}$,
as proposed in \cite{Favaro:Teh:2013}.  The efficiency of this method depends
on the sampling process and we leave such investigations to future work.
Intuitively, $q(z_{nk}|U_{n-1}) = 0$ for some $k$ when $\sum_{i = 1}^{n-1}
\hat{q}_i(z_{ik}) < \sigma$ since $q(\pi_k|U_{n-1})$ will be degenerate in
Eq.~\eqref{eqn:pik}.  This means that $\sigma$ acts as a natural threshold for
the instantiated clusters as clusters with mass (under the variational
distribution) smaller than $\sigma$ will have zero probability of having
observations assigned to it.


%% file: supplementEP.tex
\section{EP-NRM derivation}

In this section we modify the EP derivation in~\cite{Minka:2001} for our EP-NRM
algorithm for batch inference.  The resulting algorithm is conceptually similar
to ADF-NRM, except now we also save a local contribution to the variational
approximation for each data point.  The algorithm cycles through the
observations repeatedly, refining
the variational approximations for $z_{1:N}$ and $\theta$. Due to the fact that
local contributions must be saved, the algorithm is applicable to moderately
sized data sets. The full EP-NRM algorithm is shown in Alg.~\ref{alg:ep}.

Assume we have an approximation to the batch posterior
\begin{align}
p(\theta, z_{1:N}|x_{1:N}) &\approx \hat{q}(\theta, z_{1:N}) \\
&= \prod_{k = 1}^{\infty} \hat{q}(\theta_k) \prod_{i = 1}^N \hat{q}(z_i)\\
 &\propto \prod_{i = 1}^{N} \bar{q}_i(\theta,z_{1:n}),
\end{align}
where $\bar{q}_i(\theta,z_{1:n})$ are the \textit{local contributions} as
described in the main text. Furthermore, assume that
\begin{align} \label{eqn:bar}
\bar{q}_i(\theta, z_{1:n}) &= \bar{q}_i(z_i) \prod_{k = 1}^{\infty} \bar{q}_i(\theta_k)
\end{align}
and that $\bar{q}_i(z_i) = \hat{q}(z_i)$. This holds initially since since
$\bar{q}_i(\theta, z_{1:n})$ is initialized during ADF to
$\bar{q}_i(\theta, z_{1:n}) \propto
\frac{\hat{q}_i(\theta, z_{1:i})}{\hat{q}_{i-1 }(\theta,z_{1:i-1})}$.  Since
$\bar{q}_i(\theta,z_{1:N})$ only depends on $z_i$ we henceforth refer to this
quantity as $\bar{q}_i(\theta,z_i)$.  Under these assumptions we can rewrite
the approximation to the full posterior excluding data point $i$ as
\begin{align}
\hat{q}_{\setminus i}(\theta,z_{\setminus i}) &\propto \frac{\hat{q}(\theta, z_{1:N})}{\bar{q}_i(\theta,z_i)} \\
&= \frac{ \prod_{k = 1}^{\infty} \hat{q}(\theta_k) \prod_{j = 1}^n \hat{q}(z_j)}{\hat{q}(z_i) \prod_{k = 1}^{\infty} \bar{q}_i(\theta_k)} \\
&= \prod_{k = 1}^{\infty} \frac{\hat{q}(\theta_k)}{\bar{q}_i(\theta_k)} \prod_{j \neq i} \hat{q}(z_j). \label{eqn::minusup}\\
&= \prod_{k = 1}^{\infty} \hat{q}_{\setminus i}(\theta_k) \prod_{j \neq i} \hat{q}(z_j).
\end{align}
The EP-NRM algorithm consists of the following two steps. First, update the
global variational approximations, $\hat{q}(\theta_k)$ and $\hat{q}(z_i)$.
Second, use these to refine 
$\bar{q}_i(\theta_k)$ and $\bar{q}_i(z_i)$ (see Alg. \ref{alg:ep}). The global variational approximations are themselves updated using the two step procedure specified in Section 3 of the main text. Specifically, we first form $\hat{p}(z_{1:N}, \theta)|x_{\setminus i}) \stackrel{\triangle}{\propto} p(z_i|z_{\setminus i}) \hat{q}_{\setminus i}(\theta,z_{\setminus i})$, and solve
\begin{equation}
q^{\text{pr}}(z_{1:n}, \theta) = \argmin_{q \in \mathcal{Q}} \KL \Big ( \hat{p}(z_{1:N},\theta | x_{\setminus i}) || q(z_{1:n}, \theta) \Big ).
\end{equation}
We then form $\hat{p}(z_{1:N}, \theta)|x_{1:N}) \stackrel{\triangle}{\propto} p(x_i|z_i,\theta) q^{\text{pr}}(z_{1:n}, \theta)$ and solve
\begin{equation}
\hat{q}(z_{1:N}, \theta) = \argmin_{q \in \mathcal{Q}} \KL \Big ( \hat{p}(z_{1:N},\theta | x_{1:N}) || q(z_{1:n}, \theta) \Big ).
\end{equation} 
As in ADF-NRM $\hat{q}(z_j)$ for $j \neq i$ terms are unchanged; the optimal
update for $\hat{q}(z_i)$ is given by:
\begin{equation} \label{local:ep}
\hat{q}(z_{ik}) \propto q^{\pred}(z_{ik}) \int p(x_i|z_{ik},
\theta_k) \hat{q}(\theta_k) d \theta_k  \;\;\; k = 1, \ldots, K+1,
\end{equation}
where $K$ is the number of instantiated clusters and $\hat{q}(\theta_{K + 1}) = p(\theta_{K+1})$.  Similar to ADF, the predictive distribution for the NGGP in Eq.\eqref{local:ep} is given by:
\begin{equation}
   q^{\pred}_{\setminus i} (z_{ik}) {\propto} 
\begin{cases}
\displaystyle \max \left( \sum_{i \neq j} \hat{q}(z_{jk}) - \sigma, 0 \right), &k \leq K \\
a(\hat{U}_{\setminus i} + \tau)^{\sigma}, &k = K+1
\end{cases}
\end{equation}
where $q(U_{\setminus i})$ is given by:
\begin{align}
q(U_{\setminus i}) \propto \frac{U_{\setminus i}^{N-1}}{(U_{\setminus i} + \tau)^{N - 1 - a \E_{\hat{q}(z_{\setminus i})}[K']}} e^{-\frac{a}{\sigma}(U_{\setminus i} + \tau)^{\sigma}}.
\end{align}
where $K'$ is the number of unique assignments in $z_{\setminus i}$ and
$\hat{U}_{\setminus i} = \arg \max q(U_{\setminus i})$. 
 
Following the ADF discussion in the main text, the optimal variational distributions for the $\theta_k$s are given by:
 \begin{align}
 \hat{q}(\theta_k) \propto p(x_i|z_{ik},\theta_k)^{\hat{q}(z_{ik})} \hat{q}_{\setminus i}(\theta_k).
 \end{align}
 Given updated approximations $\hat{q}(\theta_k)$ and $\hat{q}(z_i)$, the local
 contribution for observation $i$ is refined as:
 \begin{align}
 \label{eqn:localcontrib}
 \bar{q}_i(\theta, z_{i}) &= \frac{\hat{q}(\theta,z_{1:N})}{\hat{q}_{\setminus i}(\theta, z_{\setminus i})} \\
 &=  \hat{q}(z_{ik}) \prod_{k = 1} ^{\infty} p(x_i|z_{ik},\theta_k)^{\hat{q}(z_{ik})} \\
 &= \bar{q}_i(z_i)  \prod_{k = 1}^{\infty} \bar{q}_i (\theta_k)
 \end{align}
which takes the form we assumed in Eq. \eqref{eqn:bar}.

When $\hat{q}(\theta_k)$ is in the exponential family with sufficient
statistics $\hat{\nu}^k$, then Eqs.~\eqref{eqn::minusup} and
\eqref{eqn:localcontrib} are given adding and subtracting the corresponding
sufficient statistics~\cite{Minka:2001}.
 
\begin{algorithm}
 \caption{EP-NRM algorithm} 
 \begin{algorithmic} \label{alg:ep}
 \STATE {$\hat{q}(\theta_{1:K}), S_{1:K}, \bar{q}(z_{1:N}), \bar{q}_{1:N}(\theta_{1:K}) \leftarrow$ ADF-NRM$(x_{1:N})$  // Initialize via ADF with data contributions.}
 \WHILE{$\hat{q}(\theta_{1:K})$ not converged}
 	\FOR{$i = 1$ \TO $N$}
 		\STATE {$\hat{U}_{\setminus i} = \argmax q(U_{\setminus i}) $}
		\FOR{$k = 1$ to K}
            \STATE $S_k = S_k - \bar{q}_i(z_{ik})$
            \STATE $q^{\pred}(z_{ik}) \propto \max(S_k- \sigma,0)$
			\STATE $\hat{q}_{\setminus i}(\theta_k) \propto \frac{\hat{q}(\theta_k)}{\bar{q}_{i}(\theta_k)}$
			\STATE $\hat{q}(z_{ik}) \propto q^{\pred}(z_{ik}) \int p(x_i|z_{ik},\theta_k) \hat{q}_{\setminus i}(\theta_k) d \theta_k$.
		\ENDFOR
		\STATE $q^{\pred}(z_{i, K + 1}) \propto a(\hat{U}_{\setminus i} + \tau)^{\sigma}$
		\STATE $\hat{q}(z_{i, K + 1}) \propto q^{\pred}(z_{i, K + 1}) \int p(x_i|z_{i, K+1},\theta) p(\theta_{K+1}) d \theta_{K+1}$
		\STATE normalize $\hat{q}(z_{i(1:K + 1)})$
		\IF {$\hat{q}(z_{i, K + 1}) > \epsilon$} 
			\STATE $K = K + 1$, $S_K = 0$, $\hat{q}(\theta_K) = p(\theta_K)$
		\ELSE 
			\STATE{normalize $\hat{q}(z_{i(1:K)})$}
		\ENDIF
		\FOR{$k = 1$ to $K$}
			\STATE $\hat{q}(\theta_k) \propto p(x_i|z_{ik},\theta_k)^{\hat{q}(z_{ik})} \hat{q}_{\setminus i}(\theta_k)$
			\STATE $S_k = S_k + \hat{q}(z_{ik})$
            \STATE $\bar{q}_i(z_{ik}) = \hat{q}(z_{ik})$
			\STATE $\bar{q}_i(\theta_k) \propto p(x_i|z_{ik},\theta_k)^{\hat{q}(z_{ik})}$
		\ENDFOR
		\STATE {Remove all clusters for which $S_k < \epsilon$}
	\ENDFOR
 \ENDWHILE
 \end{algorithmic}
 \end{algorithm}

%% file: experiments_supp.tex
\section{Experiments}

In this section we provide details on how we select hyperparameter values of
the IG and DP for the experiments in the main text.  We also present additional
experimental results regarding the convergence of EP-NRM and comparisons with the Gibbs sampler.

\subsection{Selecting Hyperparameters: $a$, $\tau$, and $\alpha$}

In order to compare the modeling performance of the IG and DP on the document
corpora considered in the main text,
we must first select the values of the hyperparameters, $a$ and $\tau$
(since $\sigma$ is known in both cases).
It is well known that the
hyperparameters of both the IG ($a$ and $\tau$) and the DP ($a$) strongly
affect the posterior
distribution over the number of inferred clusters.  For all experiments where
the IG and DP are compared we
adapt a method to determine the hyperparameters for GGP mixture models
originally developed
for batch inference~\cite{Barrios:Lijoi:Nieto-Barajas:Prunster:2013} to the
streaming setting of interest.  Specifically, for a given corpus, we consider a
small subset of the entire corpus (5\% for the NYT corpus and 10\% for both the
KOS and synthetic data) which we then split into a training and testing sets
used to determine the hyperparameters.
We run ADF-NRM on the training portion of the subset of documents (95\% of the
subset for NYT and 80\% for both KOS and synthetic data) for a grid of
parameters $a \in [1, 10, 100, 1000]$ and $\tau \in [.1, 1, 10, 100, 1000]$.
For the DP we only consider $a$ and for the IG we consider both $a$ and $\tau$.
For each parameter value we compute the
heldout log-likelihood of the test portion of the subset and choose the
values of $a$ and $\tau$ with the
largest heldout log-likelihood to use when running ADF-NRM and EP-NRM
on the remainder of the corpora.  This setup mimics a streaming scenario in
that an initial subset of the data is collected for preliminary analysis and
then the algorithm is let loose on the entire data set as it arrives.

For the Pitman-Yor data set, the grid search resulted in $a = 100$ for the DP
and $a = 1, \tau=1000$ for the IG.  The resulting parameter values for the KOS
corpus were $a=100$ for the DP and $a=10, \tau=100$ for the IG.  Last, on the
NYT corpus we obtained the parameter values $a=1000$ for the DP and
$a=100, \tau=100$ for the IG.

In the synthetic bars experiments $\alpha$ was set to $0.5$, however correct
recovery of the bars was robust to values within a reasonable range, $\alpha
\in [0.1, 0.9]$.  For the Pitman-Yor synthetic data,
the cluster centers were drawn from a Dirichlet with $\alpha = 0.75$ to ensure
overlap between clusters; $\alpha = 0.75$ was used for inference as well. For
the KOS corpus $\alpha = 0.1$ was used because it was found to provide the best
overall fit under repeated trials.
Finally, for
the NYT data set $\alpha = 0.5$ was used, as is common for this corpus
\cite{Wang:Blei:2012}.

\subsection{KOS Corpus}

\begin{figure}
\centering
\begin{subfigure}[b]{0.3\textwidth}
	\includegraphics[width=\textwidth]{figures/kos_lik_EP.pdf}
\end{subfigure}
\begin{subfigure}[b]{0.3\textwidth}
	\includegraphics[width=\textwidth]{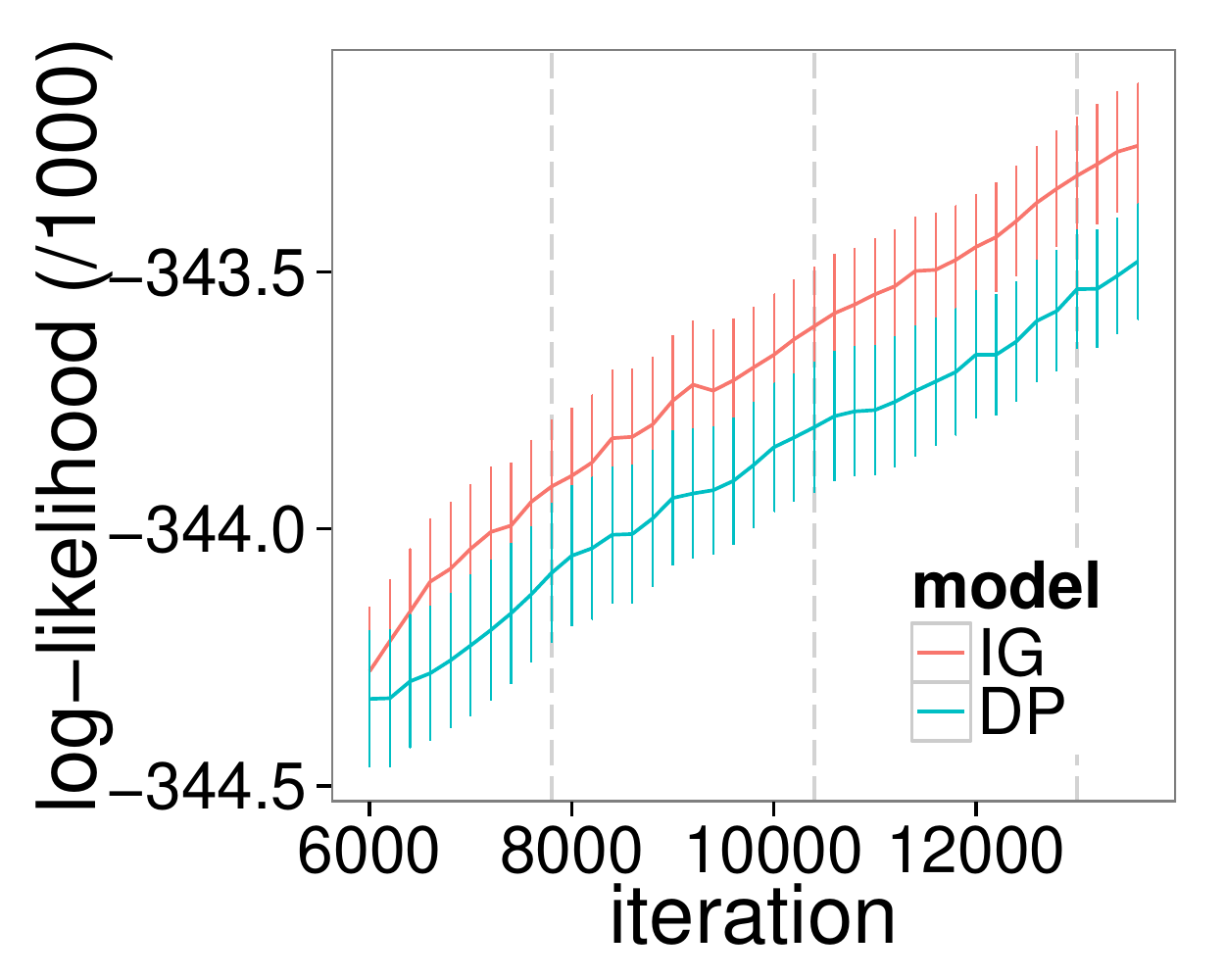}
\end{subfigure}

\caption{(\emph{left}) EP predictive performance on KOS corpus for both models
    continues to rise after the pass through the data (equivalent to ADF-NRM).
    The black vertical line indicates
the completion of the first pass through the data. The other grey vertical
lines indicate subsequent epochs.
(\emph{right}) Zoom in of the plot on the left.}

\label{fig:epplot}
\end{figure}

While the ADF-NRM algorithm makes a single pass through the corpus, a more
accurate posterior approximation can be achieved by
revisiting observations as in EP-NRM.
Figure \ref{fig:epplot} shows the predictive performance for EP-NRM applied
to the KOS corpus.  We see a rapid increase in predictive performance in the
first epoch which corresponds to ADF-NRM.  Predictive performance
continues to rise during subsequent epochs indicating an improved variational
posterior.

\begin{figure}
\centering
\includegraphics[width=.4\textwidth]{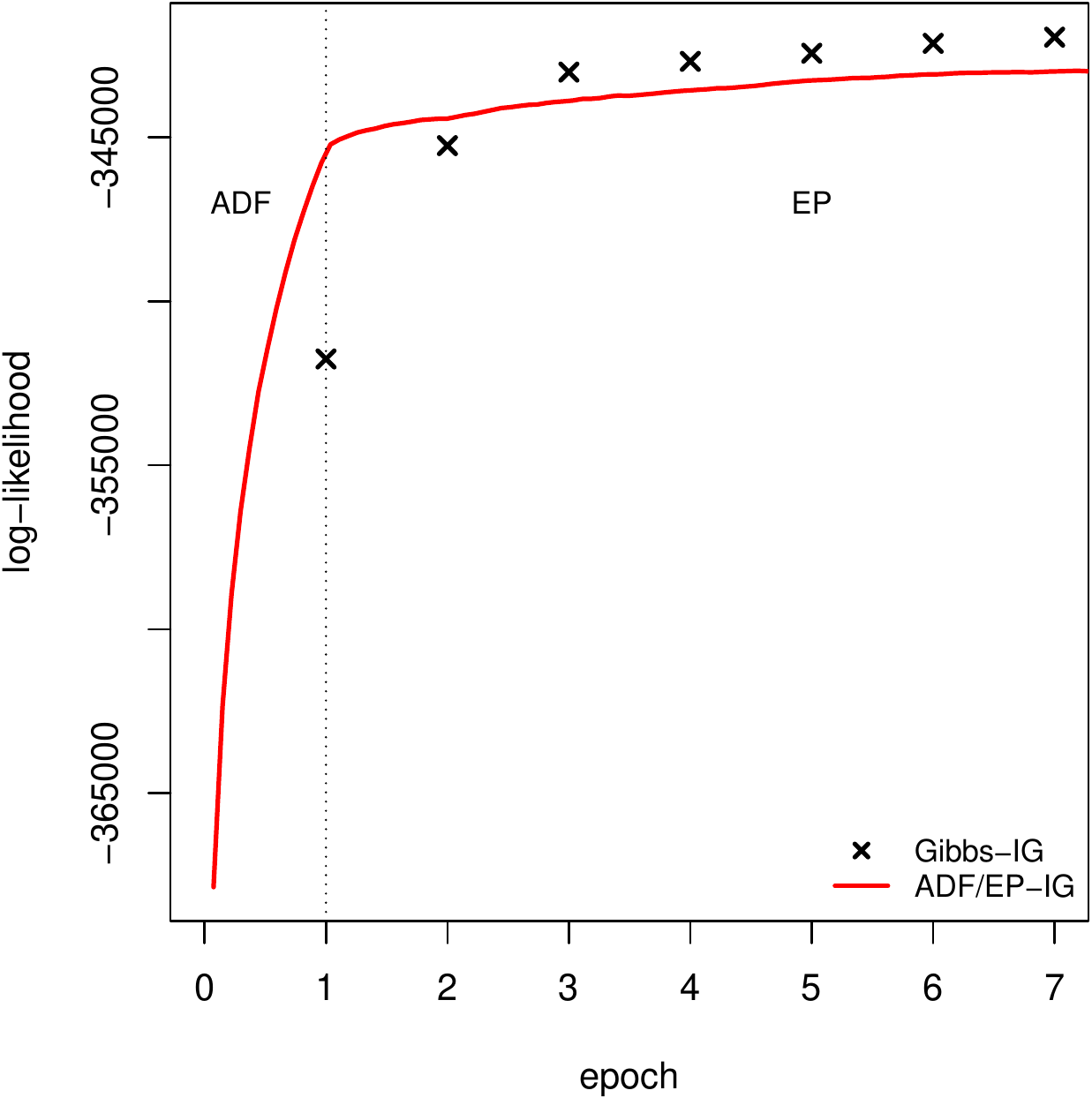}
\caption{Comparison of the predictive performance of ADF-NRM, EP-NRM, and the
    Gibbs sampler for both the IG model.  The predictive log-likelihood is
plotted against the number of epochs through the KOS corpus.}
\label{fig:epvsgibs}
\end{figure}

We compare the predictive performance of ADF-NRM, EP-NRM, and the Gibbs
sampler for the IG model on the KOS corpus and present the results in
Figure~\ref{fig:epvsgibs}.  In particular, we compare the predictive
log-likelihood of held-out data versus the number of complete passes through
the data (epochs).  Both ADF-NRM and EP-NRM are initialized as in the main text
and the Gibbs sampler is initialized so that all data points are assigned to a
single component. We found this Gibbs initialization to outperform random cluster initialization.
ADF-NRM performs significantly better than Gibbs after the first epoch and it
takes three full epochs for Gibbs to outperform ADF-NRM and EP-NRM method. Both
methods are implemented in Python and a per epoch timing comparison shows that
ADF-NRM takes an average of 220 seconds per epoch while the Gibbs sampler takes an average of 160 seconds per epoch.
The ADF-/EP-NRM methods take longer since the auxiliary variable $U$ must be
updated after each data point has been processed, while in the collapsed
sampler $U$ is only sampled once per epoch. Furthermore, in the Gibbs sampler,
after a cluster assignment has been sampled only the sufficient statistics for
the corresponding component must be updated, while in ADF-NRM, all component
parameters are updated after every data point. Importantly, our goal is not to
beat the Gibbs sampler, neither in performance nor compute time, but only to
show that the streaming ADF-NRM reaches competitive
performance to Gibbs after only a single pass through the data. Remember, Gibbs is inherently
not suited to our streaming data of interest.

\subsection{New York Times}

\begin{figure}[t!]
\centering
\includegraphics[width=0.45\textwidth]{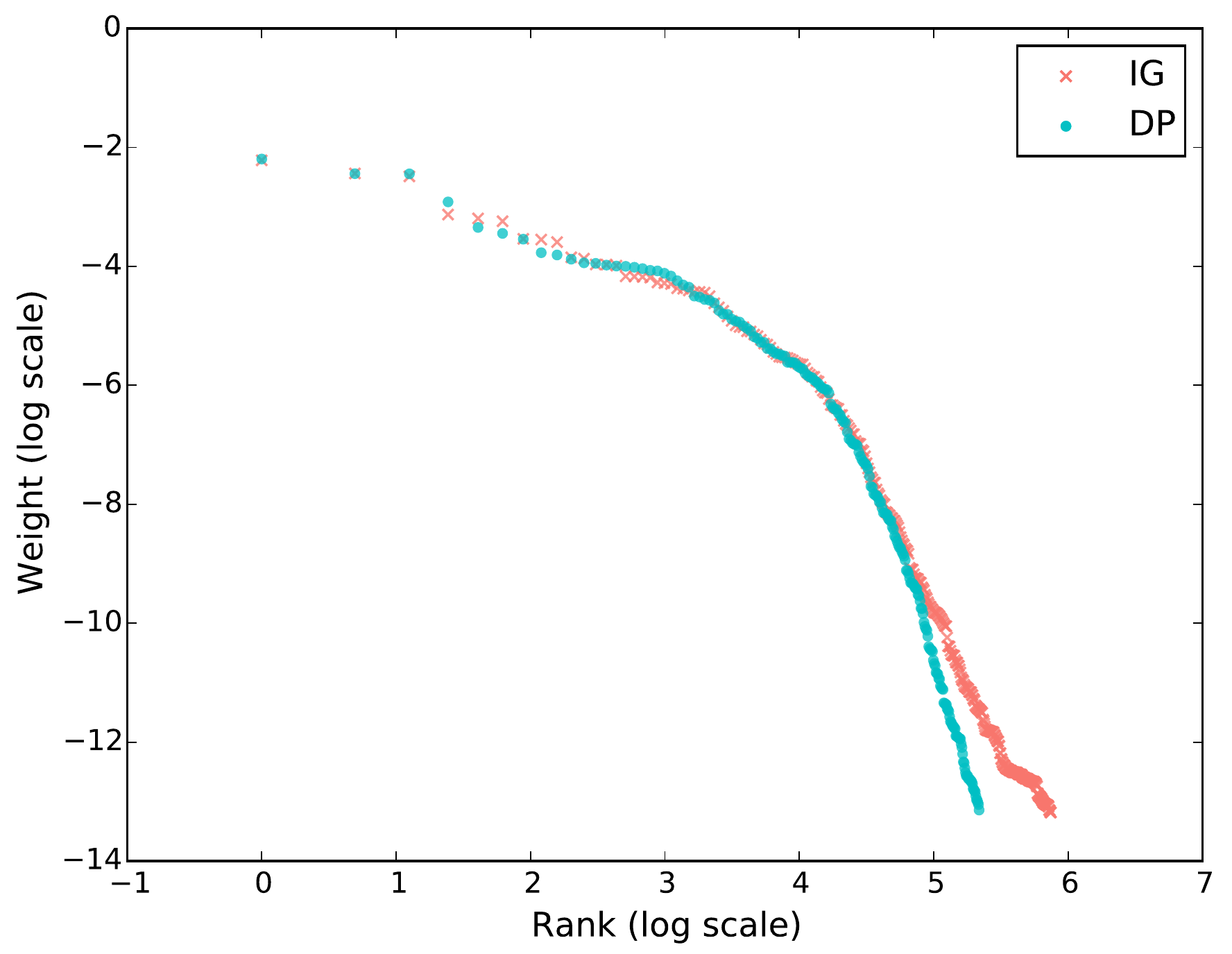}
\caption{Variational cluster weights in decreasing order.  The IG exhibits a
heavier tail than the DP.}
\label{fig:weights_loglog}
\end{figure}

As seen in the main paper, the IG both introduces more clusters than the DP
and attains superior predictive performance.  To further explore the difference
in the inferred clusters between the two models we plot the 
normalized variational cluster weights in decreasing order in
Figure~\ref{fig:weights_loglog}.  In particular,
let $S_k = \sum_{i=1}^{N} \hat{q}_i(z_{ik})$ be the total weight assigned to
cluster $k$ after a full pass through the data and
$\hat{p}_k = \frac{S_k}{\sum_{j = 1}^{K_N} S_j}$ be the
normalized weight.  We can interpret $\hat{p}_k$ as the posterior probability
of an observation being assigned to cluster $k$.
We see in Figure~\ref{fig:weights_loglog} that the distribution of weights for
the IG has a heavier tail than the DP.  The plots are similar for the large
and medium sized clusters but diverge for the small clusters, indicating that
the IG emphasizes capturing structure at a finer scale.


